\documentclass[twoside,11pt]{article}
\usepackage{amssymb} 
\usepackage{mathtools}

\newcommand{\tp}{^\top}

\newcommand{\data}{x}

\newcommand{\parameterv}{\Theta}
\newcommand{\parameterveps}{\parameterv_\radiusu}
\newcommand{\estparameterv}{\widehat{\Theta}}
\newcommand{\paraspace}{\mathcal{A}}
\newcommand{\paraspaceprime}{\paraspace'}
\newcommand{\paraspaceprimetwo}{\paraspace{''}}
\newcommand{\paramcovering}{\paraspaceprime_{\radiusu}}
\newcommand{\inputvector}{\boldsymbol{\data}}
\newcommand{\outputy}{y}
\newcommand{\datanumber}{n}
\newcommand{\noise}{u}
\newcommand{\inputdimention}{d}

\newcommand{\weightmatrix}{W}

\newcommand{\nhiddenlayer}{L}
\newcommand{\neuron}{p}

\newcommand{\R}{\mathbb{R}}

\newcommand{\networkset}{\mathcal{\functionnetbig}}
\newcommand{\networksetgeneral}{\mathcal{ \functionnetbig}_{{\generalfcn}}}

\newcommand{\functionnet}{g}
\newcommand{\functionnetbig}{G}

\newcommand{\functionclass}{\functionnet_{\parameterv}}
\newcommand{\activefunction}{f}
\newcommand{\activefunctionb}{\boldsymbol{\activefunction}}

\newcommand{\noisedis}{K}

 \newcommand{\constAtwo}{a}
\newcommand{\constA}{{a}_{\operatorname{sub}}}

\newcommand{\constk}{l}

\newcommand{\Aplus}{S}
\newcommand{\Aminus}{S}
\newcommand{\paraspaceN}{\mathcal{A}_1}

\newcommand{\LipConst}{\overline{c}_{\operatorname{Lip}}}
\newcommand{\Lipfunc}{c_{\operatorname{Lip}}(\inputvector)}
\newcommand{\LipfuncN}{c_{\operatorname{Lip}}}
\newcommand{\LipConstone}{c_{\operatorname{Lip1}}}
\newcommand{\parammetric}{\rho}

\newcommand{\wtheta}{W}
\newcommand{\wgamma}{V}
\newcommand{\actlipc}{{a}_{\operatorname{Lip}}}

\newcommand{\entropy}{ H}
\newcommand{\covernpung}{ N}

\newcommand{\allparameters}{{P}}
\newcommand{\radiusu}{r}
\newcommand{\radiuseta}{\epsilon}
\newcommand{\radiusdelta}{\delta}

\newcommand{\norm}[1]{\ensuremath{|\!| #1 |\!|_2}}
\newcommand{\normF}[1]{\ensuremath{|\!| #1 |\!|_{\operatorname{F}}}}

\newcommand{\normone}[1]{\ensuremath{\hspace{0.5mm}\!|\!| #1 | \! |_{1}}}
\newcommand{\normfcn}[1]{\ensuremath{|\!| #1 |\!|}}
\newcommand{\normP}[1]{\ensuremath{|\!|\!| #1 |\!|\!|_2}}
\newcommand{\normFP}[1]{\ensuremath{|\!|\!| #1 |\!|\!|_{\operatorname{F}}}}

\newcommand{\normoneP}[1]{\ensuremath{\hspace{0.5mm}\!|\!|\!| #1 | \! |\!|_{1}}}

\newcommand{\rbrt}[1]{(#1)} 
\newcommand{\rbrtb}[1]{\bigl(#1\bigr)}
\newcommand{\rbrtbb}[1]{\Bigl(#1\Bigr)}
\newcommand{\rbrtbbb}[1]{\biggl(#1\biggr)}

\newcommand{\sbrtbbb}[1]{\biggl[#1\biggr]}

\newcommand{\cbrt}[1]{\{#1\}} 
\newcommand{\cbrtb}[1]{\bigl\{#1\bigr\}}
\newcommand{\cbrtbb}[1]{\Bigl\{#1\Bigr\}}
\newcommand{\cbrtbbb}[1]{\biggl\{#1\biggr\}}

\newcommand{\homoconst}{a}
\newcommand{\generalfcn}{h}
\newcommand{\actfcnvar}{\boldsymbol{z}}

\newcommand{\homoorder}{k}
\newcommand{\level}{\ensuremath{t}}
\newcommand{\subparaspace}{\paraspace_{\generalfcn}}

\newcommand{\parasup}{\ensuremath{\Omega}}
\newcommand{\tuningbound}{\tuning_{\generalfcn,\level}}

\newcommand{\kapasym}{\kappa}

\newcommand{\thetasym}{\Omega}

\newcommand{\estimatedkappa}{\hat{\kapasym}}
\newcommand{\estimatedtheta}{\widehat{\thetasym}}

\newcommand{\function}{g}
\newcommand{\functionspace}{\kapasym\function_{\thetasym}}
\newcommand{\tuning}{\lambda}

\newcommand{\estimatedfunctionnew}{\estimatedkappag\function_{\estimatedthetag}}

\newcommand{\wmatrixsym}{U}

\newcommand{\uppboundfunc}{R}

\newcommand{\estimatedkappag}{\hat{\kapasym}_{\generalfcn}}
\newcommand{\estimatedthetag}{\widehat{\thetasym}_{\generalfcn}} 

\newcommand{\empprocess}{\ensuremath{z_h}}
\newcommand{\eplevel}{t}

\newcommand{\subGparaone}{\gamma}

\newcommand{\Lipfcng}{k_{\operatorname{Lip}}}

\newcommand{\networksetN}{\mathcal{G}_{1}}
\newcommand{\networksetprime}{\mathcal{G}'}

\newcommand{\entropyboundgeneral}{J(\radiusdelta,\sigma,\subparaspace)}
\newcommand{\entropyboundN}{J(\radiusdelta,\sigma,\paraspaceN)}
\newcommand{\entropyboundprime}{J(\radiusdelta,\sigma,\paraspaceprime)}
\newcommand{\E}{\mathbb{E}}
\newcommand{\rademacherZeta}{\zeta}

\newcommand{\range}{t}

\newcommand{\ConstAP}{a}
\newcommand{\SyProP}{R_{\mathcal{A}}}

\newcommand{\MCN}{b}
\newcommand{\tempConst}{\tilde{a}}

\usepackage{xcolor}
\usepackage{lineno}

\usepackage{jmlr2e}

\newif\ifmydraft
\mydrafttrue
\mydraftfalse
\ifmydraft
  \pagecolor{black!20}
\fi

\usepackage{amsmath}

\DeclareMathOperator*{\argmin}{\arg\min}

\definecolor{mygray}{gray}{0.6}

\ifmydraft
\newcommand{\MH}[1]{\color{gray}``Comment from Mahsa: #1"\color{black}}
\newcommand{\fx}[1]{{\tiny \color{blue}``FX: #1"\color{black}}}
\newcommand{\fxm}[1]{{\footnotesize \color{blue}``\text{FX: }\ensuremath{#1}"\color{black}}}
\newcommand{\MHtwo}[1]{\color{green}``MH: #1"\color{black}}
\else
\newcommand{\MH}[1]{}
\newcommand{\fx}[1]{}
\newcommand{\fxm}[1]{}
\newcommand{\MHtwo}[1]{}
\fi

\firstpageno{1}

\begin{document}

\title{Statistical Guarantees for\\Regularized Neural Networks}

\author{\name  Mahsa Taheri \email mahsa.taheri@rub.de \\
       \addr Department of Mathematics\\
  Ruhr-University Bochum\\
  Universit\"{a}tsstraße 150, 44801 Bochum, Germany
       \AND
       \name Fang Xie\email fang.xie@rub.de \\
       \addr  Department of Mathematics\\
  Ruhr-University Bochum\\
  Universit\"{a}tsstraße 150, 44801 Bochum, Germany
   \AND
       \name Johannes Lederer \email johannes.lederer@rub.de \\
       \addr  Department of Mathematics\\
  Ruhr-University Bochum\\
  Universit\"{a}tsstraße 150, 44801 Bochum, Germany}

\editor{}

\maketitle

\begin{abstract}
Neural networks have become standard tools in the analysis of data,
but they lack comprehensive mathematical theories.
For example,
there are very few statistical guarantees for learning neural networks from data, 
especially for classes of estimators that are used in practice or at least similar to such.
In this paper,
we develop a general statistical guarantee for estimators that consist of a least-squares term and a regularizer.
We then exemplify this guarantee  with $\ell_1$-regularization,
showing that the corresponding prediction error increases at most logarithmically  in the total number of parameters and can even decrease in the number of layers.
Our results establish a mathematical basis for  regularized estimation of neural networks, 
and they deepen our mathematical understanding of neural networks and deep learning more generally.
\end{abstract}


\section{Introduction}\label{sec:intro}

Neural networks have proved extremely useful across a variety of   applications,
including speech recognition~\citep{Hinton2012,Graves2013,Chorowski2015}, natural language processing~\citep{Jozefowicz2016},
object categorization~\citep{Girshick2014,Szegedy2015}, and image segmentation~\citep{Long2015,Badrinarayanan2017}.
But our mathematical understanding of neural networks and deep learning has not developed at the same speed.

A central  objective is to equip methods for learning neural networks with statistical guarantees.
Some guarantees are available for unconstrained estimators~\citep{Anthony2009},
but these 
bounds are linear in the number of parameters,
which conflicts with the large sizes of typical networks.
The focus has thus shifted to estimators that involve constraints or regularizers.
Recently surged in popularity have estimators with  $\ell_1$-regularizers~\citep{Bartlett1998,Bartlett2002,Anthony2009,Barron2018,Barron2019,Liu2019},
motivated by the success of this type of regularization in linear regression~\citep{lasso},
compressed sensing~\citep{Candes2006,Donoho2006},
and many other parts of data science.
A key feature of $\ell_1$-regularization is that it is easy to include into  optimization schemes and, at the same time, induces sparsity,
which has a number of favorable effects in deep learning~\citep{Glorot2011}.
There has been some progress on  guarantees for least-squares with  constraints based on  
the sparsity of the networks~\citep{Hieber2017} or group-type norms on the weights~\citep{Neyshabur2015}.
These developments have provided valuable intuition, for example,
about the role of network widths and depths,
but important problems remain:
for example, the  combinatorial  constraints in the first paper render the corresponding estimators infeasible in practice,
 the exponential dependence of the bounds in the second paper are contrary to the trend toward very deep networks.
More generally,
many questions about 
the statistical properties of constraint and regularized estimation of neural networks remain open.

In this paper, 
we introduce a general class of regularized least-squares estimators.
Our strategy is to disentangle the parameters into a ``scale'' and a ``direction''---similarly to introducing polar coordinates---which allows us to focus the regularization on a one-dimensional parameter.
We call our approach \emph{scale regularization}.
We then equip the  scale regularized least-squares estimators with a general statistical guarantee for prediction.
A main feature of this guarantee is that it connects neural networks to standard empirical process theory through a quantity that we  call the \emph{effective noise}.
This connection facilitates the specification of the bound to different types of regularization.

In a second step, we  exemplify the general bound for $\ell_1$-regularization.
We find a guarantee for the squared prediction error of the order of 
\begin{equation*}
(L/2)^{1/2-\nhiddenlayer}\sqrt{\log\rbrt{\allparameters}}\,\frac{\log(\datanumber)}{\sqrt{\datanumber}},
\end{equation*}
which decreases essentially as~$1/\sqrt{n}$ in the number of samples~$n$, 
increases only logarithmically in the total number of parameters~$\allparameters$,
and---everything else fixed---decreases in the number of hidden layers~$\nhiddenlayer$. 
This result suggests that $\ell_1$-regularization can ensure accurate prediction even of very wide and deep networks.

In Section~\ref{sec:mainresult}, we introduce our regularization scheme and establish a general prediction bound that allows for different types of regularization.
In Section~\ref{sec:example-l1}, we specify this  bound to $\ell_1$-regularization.
In Section~\ref{sec:techresults}, we establish Lipschitz and complexity properties of neural networks.  Section~\ref{sec:proofs},
we give detailed proofs.
In Section~\ref{sec:discussion}, we conclude our paper and discuss some limitations.

\section{Scale regularization for neural networks}\label{sec:mainresult}
We first establish an alternative parametrization of neural networks and use this parameterization to define our regularization strategy.
We then provide a prediction guarantee for the corresponding estimators.

\subsection{Alternative parametrization}
\label{subsec:alterpara}
\newcommand{\functiontrue}{\ensuremath{g_*}}
\newcommand{\functiontruei}{\ensuremath{g_*(\inputvector_i)}}
Consider data~$\rbrt{\inputvector_1,\outputy_1},\dots,\rbrt{\inputvector_\datanumber,\outputy_\datanumber}\in\R^{\inputdimention}\times\R$
 that follow a regression model
\begin{equation}
\label{model}
    \outputy_i=\functiontruei+\noise_i~~~~~~~~~~~~~~\text{for}~i\in\{1,\dots,\datanumber\}
\end{equation}
for some function~$\functiontrue\,:\,\R^{\inputdimention}\to\R$.
We are interested in approximating~\functiontrue\ based on neural networks.
Following first standard approaches, we consider feedforward neural networks of the form 
\begin{equation}
\label{nns}
\begin{aligned}
    \functionclass\ :\ \R^\inputdimention&\to\R\\
    \inputvector&\mapsto \functionclass(\inputvector):=\weightmatrix^{\nhiddenlayer}\activefunctionb^{\nhiddenlayer}\rbrtb{\dots\weightmatrix^{1}\activefunctionb^{{1}}\rbrt{\weightmatrix^{0}\inputvector}}
\end{aligned}
\end{equation}
indexed by the network parameter~$\parameterv=\rbrt{\weightmatrix^{\nhiddenlayer},\ldots,\weightmatrix^{0}}$ that summarizes the weight matrices~$\weightmatrix^{l}\in\R^{p_{l+1}\times p_l}$.
The~$\inputvector_{i}$ and~$\outputy_i$ are the network's {inputs} and {outputs}, respectively,
and the~$\noise_i$ are the {noise} variables.
For ease of notation,
the~$\inputvector_{i}$ are fixed except in the generalization bounds.
The network's architecture is specified by the {number of hidden layers} or  {depth}~$\nhiddenlayer\in\{1,2,\dots\}$ and by the the number of neurons in each layer or width~$p_0,\dots,p_{\nhiddenlayer+1}\in \{1,2,\dots\}$. 
The~$0$th layer is the {input layer} with~$\neuron_0=\inputdimention$, and the~$(\nhiddenlayer+1)$th layer is the {output layer} with~$\neuron_{\nhiddenlayer+1}=1$.
The total number of parameters is $\allparameters:=\sum_{l=0}^{\nhiddenlayer}\neuron_{l+1}\neuron_{l}$.
The functions~$\activefunctionb^l\,:\,\R^{\neuron_{l}}\to\R^{\neuron_{l}}$ are called {activation functions}.
We omit shifts in the activation functions for notational simplicity,
but such can often be incorporated as additional neurons~\citep{Barron2018}.

The  parameter space in the above formulation is
\begin{equation*}
    \paraspace:=\cbrtb{\parameterv =\rbrt{\weightmatrix^\nhiddenlayer,\ldots,\weightmatrix^0}\ :\  \weightmatrix^l\in\R^{\neuron_{l+1}\times\neuron_l}}.
\end{equation*}
In the following, however, we propose an alternative parametrization.
We say that a function\newcommand{\genfunction}{q}
$\boldsymbol{\genfunction}\,:\,\R^s\to \R^t$ is nonnegative homogeneous of degree~$\homoorder\in(0,\infty)$ if 
\begin{equation*}
    \boldsymbol{\genfunction}(\homoconst\actfcnvar)=\homoconst^\homoorder\boldsymbol{\genfunction}(\actfcnvar)~~~~~~~~~~\text{for all~}\homoconst\in[0,\infty)\text{~and~}\actfcnvar\in\R^{s}
\end{equation*}
and we say that a function~$\genfunction\,:\,\R^s\to[0,\infty)$ is positive definite if 
\begin{equation*}
\genfunction(\boldsymbol{z})=0~~~~\Leftrightarrow~~~~\boldsymbol{z}=\boldsymbol{0}_{s}. 
\end{equation*}
The corresponding properties for functions on~$\paraspace$ are defined accordingly.
For example, every norm on~$\R^s$ or~$\paraspace$ is nonnegative homogeneous of degree~1 and positive definite.
We then find the following:
\begin{proposition}[Equivalence between neural networks]
\label{prop:equNNs}
Assume that the activation functions~$\activefunctionb^1,\dots,\activefunctionb^\nhiddenlayer$ are nonnegative homogeneous of degree~$1$. 
Consider a  function~$\generalfcn\,:\, \paraspace\to[0,\infty)$
that is nonnegative homogeneous of degree~$\homoorder\in(0,\infty)$ and positive definite,
and denote the corresponding unit ball by
\begin{equation*}
    \paraspace_\generalfcn:=\bigl\{\parameterv\in\paraspace\ :\ \generalfcn(\parameterv)\leq 1\bigr\}.
\end{equation*}

Then,
for every~$\parameterv\in\paraspace$, there exists a pair of~$\kapasym\in[0,\infty)$ and~$\thetasym\in\paraspace_h$
such that 
\begin{equation*}
    \function_{\parameterv}(\inputvector)=\functionspace(\inputvector)~~~~~~~~~\text{for all~}\inputvector\in\R^{\inputdimention};
\end{equation*}
and vice versa,
for every pair of~$\kapasym\in[0,\infty)$ and~$\thetasym\in\paraspace_\generalfcn$,
there exists a~$\parameterv\in\paraspace$ 
such that the above equality holds.
\end{proposition}
\noindent Proposition~\ref{prop:equNNs} is just a formulation of the known fact that weights can be rescaled across layers that have nonnegative-homogeneous activations~\citep{Du2018,Hebiri20,Neyshabur2014}.
The interesting part of this section is not Proposition~\ref{prop:equNNs} itself but the observation that this rescaling can lead to a reparameterization that is particularly suitable for regularization.
Motivated by Proposition~\ref{prop:equNNs}, we change the parameter space for estimating the true data generating function~$\functiontrue$ to~$[0,\infty)\times\subparaspace$ and the corresponding space of networks to~$\cbrt{\functionspace\,:\,\kapasym\in[0,\infty),\thetasym\in\subparaspace}$.
In other words, we study the neural networks 
\begin{equation}
\label{nnsmod}
\begin{aligned}
    \functionspace\ :\ \R^\inputdimention&\to\R\\
    \inputvector&\mapsto \functionspace(\inputvector):=\kapasym\wmatrixsym^{\nhiddenlayer}\activefunctionb^{\nhiddenlayer}\rbrtb{\dots\wmatrixsym^{1}\activefunctionb^{{1}}\rbrt{\wmatrixsym^{0}\inputvector}}
\end{aligned}
\end{equation}
indexed by the parameters~$\kapasym\in[0,\infty)$ and~$\thetasym=\rbrt{\wmatrixsym^\nhiddenlayer,\ldots,\wmatrixsym^0}\in\subparaspace$.
We can interpret~$\kapasym$ as the network's ``scale''  and~$\thetasym$ as the network's ``orientation.''
Proposition~\ref{prop:equNNs} ensures equivalence to  the original set of networks if the activations are nonnegative homogeneous (ReLU activations are popular examples),
but we can use the proposed parametrization more generally.
We now argue that the scale parameter is particularly suitable for regularizing the ``overall size'' of the network and the orientation parameter for specifying the desired ``type'' of the network.
In particular, 
rather than naively transferring standard regularization schemes from other parts of machine learning,
we propose to tailor these regularization schemes to the characteristics of neural networks as captured by the above parameterization.
We detail this argument in the following sections,
where we introduce concrete regularization schemes and develop statistical guarantees.
These statistical guarantees are the main result of this paper.

\subsection{Estimation}
\label{subsec:estimation}
The most basic approach to fit the model parameters of the network~\eqref{nns} to the model~\eqref{model} is the least-squares estimator
\begin{equation*}
    \estparameterv_{\operatorname{LS}}\in \argmin_{\Theta\in\paraspace}\,\cbrtbbb{\frac{1}{\datanumber}\sum_{i=1}^{\datanumber}{\rbrtb{\outputy_i-\functionnet_{\Theta}(\inputvector_i)}}^2}.
\end{equation*}
But to account for the high-dimensionality of the parameter space~$\paraspace$, 
the least-squares estimator is often complemented with a regularizer~$\generalfcn\,:\,\paraspace\to[0,\infty)$;
popular choices for~$\generalfcn$ are the $\ell_1$-norm~\citep{Zhang2016L1} or group versions of it~\citep{Scardapane2017Group}.
A straightforward way to incorporate such regularizers is  
\begin{equation*}
    \estparameterv_{\operatorname{reg},\generalfcn}\in \argmin_{\Theta\in\paraspace}\,\cbrtbbb{\frac{1}{\datanumber}\sum_{i=1}^{\datanumber}{\rbrtb{\outputy_i-\functionnet_{\Theta}(\inputvector_i)}}^2+\tuning\generalfcn(\parameterv)},
\end{equation*}
where~$\tuning\in[0,\infty)$ is a tuning parameter.
But in neural network frameworks, 
it turns out difficult to analyze such estimators statistically.

We introduce, therefore, a different way to incorporate regularizers.
The approach is based on our new parametrization.
The equivalent of the above least-squares estimator in the framework~\eqref{nnsmod} is 
\begin{equation*}
    (\estimatedkappa_{\operatorname{LS}},
   \estimatedtheta_{\operatorname{LS}})\in \argmin_{\substack{\kapasym\in[0,\infty)\\\thetasym\in\subparaspace}}\,\cbrtbbb{\frac{1}{\datanumber}\sum_{i=1}^{\datanumber}{\rbrtb{\outputy_i-\functionspace(\inputvector_i)}}^2}.
\end{equation*}
It holds that~$\functionnet_{\estparameterv_{\operatorname{LS}}}=\estimatedkappa_{\operatorname{LS}} \functionnet_{\estimatedtheta_{\operatorname{LS}}}$ under the conditions of Proposition~\ref{prop:equNNs},
but we can take this estimator as a starting point more generally.
This allows us to focus the regularization on the scale-parameter~$\kapasym$;
in other words, we propose the estimators 
\begin{equation}
\label{estimator}
   (\estimatedkappag,
   \estimatedthetag)\in \argmin_{\substack{\kapasym\in[0,\infty)\\\thetasym\in\subparaspace}}\,\cbrtbbb{\frac{1}{\datanumber}\sum_{i=1}^{\datanumber}{\rbrtb{\outputy_i-\functionspace(\inputvector_i)}}^2+\tuning\kapasym},
\end{equation}
where~$\tuning\in[0,\infty)$ is a tuning parameter. 
The fixed constraint~$\thetasym\in\subparaspace$ captures the type of regularization (such as~$\ell_1$), 
while the actual regularization concerns only on the scale~$\kapasym\in[0,\infty)$.
We thus call our approach \emph{scale regularization}.

The concentration of the regularization on a one-dimensional parameter greatly facilitates the statistical analysis.
Specifically,
it will allow us to focus our attention on
\begin{equation}
    \empprocess:=\sup_{\thetasym\in\subparaspace} \Bigl|\frac{2}{\datanumber}\sum_{i=1}^{\datanumber}\functionnet_{\thetasym}(\inputvector_i) \noise_i\Bigr|.
\end{equation}
This quantity is related to the Gaussian and Rademacher complexities of the function class $\cbrt{\functionnet_{\thetasym}\,:\,\thetasym\in\subparaspace}$.
For example,
the expectation of $\empprocess$ is the Gaussian complexity of the function class $\cbrt{\functionnet_{\thetasym}\,:\,\thetasym\in\subparaspace}$ if the $\noise_i$'s are i.i.d.~standard normal random variables---cf.~\citep{Bartlett2002}, for example.
But while the Gaussian and Rademacher complexities are  standard measures for function classes,
there are two important subtleties here:
first, Gaussian and Rademacher complexities  require the specification of a distribution over the data,
which we can avoid at this point;
second,
the function class at hand does not contain the entire networks~$\functionspace$ but only their ``orientation parts''~$\functionnet_{\thetasym}$.
Therefore, we should rather think of~$\empprocess$ as the neural-network equivalent of what high-dimensional linear regression refers to as the \emph{effective noise}~\citep{Lederer2020Estimating}.

We need to ensure---just as in high-dimensional linear regression---that the effective noise is controlled by the tuning parameter with high probability.
In this spirit, we define quantiles~$\tuningbound$ of the effective noise for given level~$\eplevel\in[0,1]$ through 
\begin{equation}\label{tunbound}
    \tuningbound\in\min\bigl\{\radiusdelta\in[0,\infty)\ : \ \mathbb{P}\bigl(\empprocess\leq  \radiusdelta)\geq 1- \eplevel\bigr\}.
\end{equation}
In other words,~$\tuningbound$ is the smallest tuning parameter that controls the effective noise~\empprocess\ at level~$1-\eplevel$.

To measure the accuracy of the regularized estimators, we consider the (in-sample-) prediction error (also called ``denoising error'') with respect to the data generating function~$\functiontrue$:
\newcommand{\prederrors}[1]{\operatorname{err}^2{\!(#1)}}
\newcommand{\prederror}[1]{\operatorname{err}{\!(#1)}}
\begin{equation}\label{eq:prederrors}
   \prederror{\functionspace}:=\sqrt{\frac{1}{\datanumber}\sum_{i=1}^{\datanumber}\rbrtb{\functionspace(\inputvector_i)-\functiontrue(\inputvector_i)}^2}~~~~~~~~~~~\text{for}~\kapasym\in[0,\infty),\thetasym\in\subparaspace.
\end{equation}
This is a standard measure of how well the data generating function is learned.
An interesting feature of the in-sample-prediction error is that it avoids any distributional assumptions on the data.
Moreover,
it also entails bounds on the generalization error (also called ``out-of-sample-prediction error'' or ``prediction risk'') for a new sample $(\inputvector,\outputy)\in\R^{\inputdimention}\times\R$ 
\begin{equation*}
\operatorname{risk}(\functionspace):=\E_{(\inputvector,\outputy)}\Bigl[\bigl(\functionspace(\inputvector)-\outputy\bigr)^2\Bigr]~~~~~~~~~~~\text{for}~\kapasym\in[0,\infty),\thetasym\in\subparaspace\,,  
\end{equation*}
which is more common in the deep-learning literature---see Lemma~\ref{inSamVsOutSam}.

We find the following guarantee:
\begin{theorem}[Prediction guarantee]\label{GPbound}
Assume that~$\tuning\geq \tuningbound$ for a~$\level\in[0,1]$.
Then,
\begin{equation*}
   \prederrors{\estimatedfunctionnew}\leq \inf_{\substack{\kapasym\in[0,\infty)\\\thetasym\in\subparaspace}}\bigl\{\prederrors{\kapasym \functionnet_{\thetasym}}
   +2\tuning\kapasym\bigr\}
\end{equation*}
with probability at least~$1-\level$. 
\end{theorem}
\noindent The bound is an analog of what has been called sparsity-bound in high-dimensional linear regression~\citep{Lederer2019}.
For neural networks,
however, it is the first such bound.
It states that the squared prediction error of the regularized estimator is governed by an approximation error or squared bias~$\prederrors{\kapasym \functionnet_{\thetasym}}$ and an excess error or variance~$2\tuning\kapasym$.
In other words,
the estimator is guaranteed to have a small prediction error if (i)~the quantile~$\tuningbound$ is small and (ii)~the data generating function can be represented well by a neural network with reasonably small~$\kapasym$.
A typical example for~(i) is provided in the following section;
recent results on approximation theory support~(ii) especially for wide and deep networks~\citep{Yarotsky2017}.

Since~$\empprocess$ is a supremum over an empirical process,
it allows us to connect our statistical  theories with theories on empirical processes.
Deviation inequalities that bound quantities such as~$\tuningbound$ have been established even for noise~$\noise_i$ that has very heavy tails~\citep{Lederer2014}.
In Section~\ref{sec:example-l1}, we derive an explicit bound for~$\tuningbound$ for $\ell_1$-regularization and sub-Gaussian noise.
Crucial in this derivation, and in controlling~$\empprocess$ in general,
is that the index set of the empirical process is the constraint parameter space~$\subparaspace$ rather than the entire parameter space~$\paraspace$.
This key feature of~$\empprocess$ is due to our novel way of regularizing.

The standard parametrization~$\parameterv\in\paraspace$ of neural networks is  ambiguous:
there are typically uncountably many parameters~$\parameterv\in\paraspace$ that yield the same network~$\functionclass$.
This ambiguity remains in our new framework with~$(\kapasym,\thetasym)\in[0,\infty)\times\subparaspace$.
But importantly, 
our guarantees hold for \emph{every} solution~$(\estimatedkappag,\estimatedthetag)$.

\section[An example: l1-regularization]{An example: $\ell_1$-regularization}\label{sec:example-l1}
In view of its long-standing tradition in other parts of statistics and machine learning, 
sparsity-inducing regularization with $\ell_1$-terms has already sparked some theoretical research.
This existing research has two components:
first,  general risk bounds in terms of the fat-shattering dimension  or the Rademacher complexity such as  \citet[Theorem~2]{Bartlett1998} and \citet[Theorem~8]{Bartlett2002}, respectively;
and second, 
bounds for the fat-shattering dimension and Rademacher complexity of classes of $\ell_1$-constraint neural networks such as \citet[Section~IV.B]{Bartlett1998} and \citet{Golowich,Neyshabur2015}, respectively.
But these results have severe limitations:
for example, they require bounded losses (which excludes the least-squares loss, for example);
they consider constraints rather than regularization terms (which is the version used in practice);
they do not provide insights into how the tuning parameters should scale with the dimensions of the problem, such as the sample size, the network size, and so forth (which can eventually lead to practical advise);
and they have---except for \citet{Golowich}---a strong dependence on the network depth (which contradicts the current trend toward deep learning).

It turns out that our general theory applied to $\ell_1$-regularization can do away with these limitations.
We define~$\generalfcn$ as 
\begin{equation*}
    \generalfcn(\thetasym):=\normoneP{\thetasym}:=\sum_{l=0}^{\nhiddenlayer}\sum_{k=1}^{\neuron_{l+1}}\sum_{j=1}^{\neuron_{l}}|\wmatrixsym_{kj}^l|.
\end{equation*}
And to fix ideas,
we impose two assumptions on the activation functions and the noise:
First, we assume that the  activation functions satisfy~$\activefunctionb^l(\boldsymbol{0}_{p_l})=\boldsymbol{0}_{p_l}$ and are~$\actlipc$-Lipschitz continuous for  a constant~$\actlipc\in[0,\infty)$ and with respect to the Euclidean norms on their input and output spaces:
\begin{equation*}
    \norm{\activefunctionb^l(\boldsymbol{z})-\activefunctionb^l(\boldsymbol{z}')}\leq \actlipc\norm{\boldsymbol{z}-\boldsymbol{z}'}~~~~~~~\text{for all}~\boldsymbol{z},\boldsymbol{z}'\in \R^{\neuron_{\constk}}.
\end{equation*}
This assumption is satisfied by many popular activation functions:
for example, the coordinates of the activation functions could be  ReLU functions~$x\mapsto 0\vee x$~\citep{Nair2010Rectified}, ``leaky" versions of ReLU
~$x\mapsto (0\vee x)+(0\wedge c x)$ for~$c\in(0,1)$, 
ELU functions~$x\mapsto x\vee 0 +c(e^{x \wedge 0}-1)$ for~$c\in(0,1]$~\citep{Clevert2015ELU},
hyperbolic tangent functions~$x\mapsto (e^{2x}-1)/(e^{2x}+1)$, or SiL/Swish functions~$x\mapsto x/(1+e^{-x})$ \citep{Ramachandran2017Swish,Elfwing2018}
(throughout, we use the shorthands~$r \vee s:=\max\{r,s\}$ and~$r \wedge s:=\min\{r,s\}$ for~$r,s\in\R$).
Feasible Lipschitz constants for  these examples are~$\actlipc=1.1$ for SiL/Swish and~$\actlipc=1$ for all other functions.

Second, we assume that the 
noise variables~$\noise_i$ are independent, centered, and uniformly sub-Gaussian for  constants~$K,\subGparaone\in(0,\infty)$~(\citealp[Page~126]{Sara2000}; \citealp[Section~2.5]{Vershynin2018}):\label{subgauss}
\begin{equation*}\label{sgnoise}
    \max_{i\in\{1,\dots,\datanumber\}}\noisedis^2(\mathbb{E}e^{\frac{|\noise_i|^2}{\noisedis^2}}-1)\leq \subGparaone^2.
\end{equation*}

Using the shorthands~$
    \paraspace_1:=\cbrt{\Theta\in\paraspace\,:\,\normoneP{\parameterv}\leq 1}$ and $\normfcn{\boldsymbol{x}}_{\datanumber}:=\sqrt{\sum_{i=1}^{\datanumber}{\norm{\inputvector_i}^2}/\datanumber}$, we then find the following prediction guarantee for the estimator in~\eqref{estimator}:
\begin{theorem}[Prediction guarantee for $\ell_1$-regularization]\label{UPBOUA1}
Assume that 
\begin{equation*}
  \tuning\geq \constAtwo\Bigl(\frac{2\actlipc}{L}\Bigr)^\nhiddenlayer\normfcn{\inputvector}_{\datanumber}  \sqrt{\nhiddenlayer\log(2\allparameters)}\,\frac{\log(2\datanumber)}{\sqrt{\datanumber}},
\end{equation*}
where~$\constAtwo\in(0,\infty)$ is a constant that depends only on the sub-Gaussian parameters~$\noisedis$ and~$\subGparaone$ of the noise. 
Then,
for~$\datanumber$ large enough,
\begin{equation*}
   \prederrors{\estimatedfunctionnew}\leq
   \inf_{\substack{\kapasym\in[0,\infty)\\\thetasym\in\paraspaceN}}\bigl\{\prederrors{\kapasym \functionnet_{\thetasym}}
   +2\tuning\kapasym\bigr\}
\end{equation*}
 with probability at least~$1-1/\datanumber$.
\end{theorem}

\noindent 
The bound establishes essentially a~$1/\sqrt{n}$-decrease of the error in the sample size~$n$, 
a mild logarithmic increase in the number of parameters~$\allparameters$,
and an almost exponential decrease in the number of hidden layers~$\nhiddenlayer$ if everything else is fixed (for example, the number of parameters~\allparameters\ can depend on the number of hidden layers~$\nhiddenlayer$).
The dependencies on the sample size~$n$ and the number of parameters~$\allparameters$ match those of standard bounds in $\ell_1$-regularized linear regression~\citep{Hebiri2013Correlations}.
But one can argue that the logarithmic dependence on the number of parameters is even more crucial for neural networks: 
already a small network with~$\nhiddenlayer=10$,~$\neuron_0=100$, and~$\neuron_1,\dots,\neuron_\nhiddenlayer=50$ 
involves~$\allparameters=27\,550$ parameters,
which highlights that neural networks typically involve very large~$\allparameters$.

As an illustration,
we can simplify Theorem~\ref{UPBOUA1} further in a parametric setting:
\newcommand{\kapastarone}{\kappa_*}
\newcommand{\thetastarone}{\Omega_*}
\begin{corollary}[Parametric setting]\label{Paramset}
Assume that
\begin{equation*}
\tuning=\constAtwo\Bigl(\frac{2\actlipc}{L}\Bigr)^\nhiddenlayer\normfcn{\inputvector}_{\datanumber}  \sqrt{\nhiddenlayer\log(2\allparameters)}\,\frac{\log(2\datanumber)}{\sqrt{\datanumber}}
\end{equation*}  
and that there exist  parameters~$(\kapastarone,\thetastarone)\in [0,\infty)\times\paraspaceN$ such that~$\kapastarone\functionnet_{\thetastarone}(\inputvector_i)=\functiontrue(\inputvector_i)$ for all~$i\in\cbrt{1,\dots,\datanumber}$.
Then, for~$\datanumber$ large enough,
\begin{equation*}
   \prederrors{\estimatedfunctionnew}\leq 2\constAtwo\kapastarone\Bigl(\frac{2\actlipc}{L}\Bigr)^\nhiddenlayer\normfcn{\inputvector}_{\datanumber}  \sqrt{\nhiddenlayer\log(2\allparameters)}\frac{\log(2\datanumber)}{\sqrt{\datanumber}}
\end{equation*}
with probability at least~$1-1/2\datanumber$.
\end{corollary}

The above choice of~$\generalfcn$ is not the only way to formulate $\ell_1$-constraints.
Another way is, for example,~$\generalfcn(\thetasym):=\max_{l\in\{0,\dots,\nhiddenlayer\}}\sum_{k=1}^{\neuron_{l+1}}\sum_{j=1}^{\neuron_{l}}|\wmatrixsym_{kj}^l|$.
The proofs and results remain virtually the same,
and one may choose in practice whatever  regularizer is more appropriate or easier to compute.
And more broadly,
our theories provide a general scheme for deriving prediction guarantees that could account for different regularizers (such as grouped versions of~$\ell_1$), 
activation functions (such as non-Lipschitz functions),
and noise (such as heavy-tailed noise) through corresponding bounds for~$\empprocess$.

The bounds in the in-sample-prediction error also entail bounds in the generalization error.
We illustrate this here in the case of $\ell_1$-regularization.
We assume that the input data are random and find:
\begin{lemma}[Generalization guarantee for $\ell_1$-regularization]\label{inSamVsOutSam}
Assume that the  conditions of Corollary~\ref{Paramset} are satisfied and that the inputs $\inputvector_1,\dots,\inputvector_{\datanumber}$ are independent random vectors. 
Then, for~$\datanumber$ large enough,
\begin{multline*}
\operatorname{risk}(\estimatedfunctionnew)
\leq 1.01\,\operatorname{risk}(\kapastarone\functionnet_{\thetastarone}) 
+\constAtwo\kapastarone\Bigl(\frac{2\actlipc}{L}\Bigr)^\nhiddenlayer\normfcn{\inputvector}_{\datanumber}  \sqrt{\nhiddenlayer\log(2\allparameters)}\,\frac{\log(2\datanumber)}{\sqrt{\datanumber}}\\
+\ConstAP(\kapastarone)^2\Bigl(\frac{2\actlipc}{L}\Bigr)^{2\nhiddenlayer}\sqrt{\nhiddenlayer^2\log(2\allparameters)\sum_{i=1}^{\datanumber}\norm{\inputvector_i}^4}\,\frac{\log(2\datanumber)}{\datanumber}
\end{multline*}
with probability at least~$1-1/\datanumber$, where~$\ConstAP\in(0,\infty)$ is a constant that depends only on the sub-Gaussian parameters~$\noisedis$ and~$\subGparaone$ of the noise.
\end{lemma}
\noindent The result ensures that the estimator approaches the oracle risk at the above-discussed rate.

\section{Further technical results}
\label{sec:techresults}
We now establish  Lipschitz and complexity properties of neural networks.
These results are used in our proofs but might also be of interest by themselves.
To start, we define operator norms of the parameters and the weight matrices by
\begin{equation*}\label{paramnormtwo}
 \normP{\parameterv}:=\sqrt{\sum_{l=0}^{\nhiddenlayer}\norm{\weightmatrix^{l}}^2}\ \ \ \ {\rm and}\ \ \ \  \norm{\weightmatrix^l}:=\sigma_{\max}(\weightmatrix^l),
\end{equation*}
respectively, where~$\sigma_{\max}(\weightmatrix^l)$ is the largest singular value of~$\weightmatrix^l$.
We also define Frobenius norms of the parameter and weight matrices by
\begin{equation*}\label{Fnormmatrix}
  \normFP{\parameterv}:=\sqrt{\sum_{l=0}^{\nhiddenlayer} \normF{\weightmatrix^l}^2}\ \ \ \ {\rm and}\ \ \ \  \normF{\weightmatrix^l}:= \sqrt{\sum_{k=1}^{p_{l+1}} \sum_{j=1}^{p_{l}} \rbrt{\weightmatrix^l_{kj}}^2}.
\end{equation*}
We then define the Euclidean norm of vectors by~$\norm{\boldsymbol{v}}:={\sqrt{\sum_{i=1}^{\inputdimention}\rbrt{v_i}^2}}$ for~$\boldsymbol{v}\in\R^{\inputdimention}$.
And finally, the prediction distance of any two networks~$\functionnet_{\parameterv}$ and~$\functionnet_{\Gamma}$ with~$\parameterv,\Gamma\in\paraspace$ is 
\begin{equation*}\label{empnorm}
   \normfcn{{\functionnet_{\parameterv}-\functionnet_{{\Gamma}}}}_{\datanumber}:=\sqrt{\frac{1}{\datanumber}\sum_{i=1}^{\datanumber}\rbrtb{\functionnet_{\parameterv}(\inputvector_i)-\functionnet_{\Gamma}(\inputvector_i)}^2},
\end{equation*}
and similarly,
\begin{equation*}
    \normfcn{{\functionnet_{\parameterv}}}_{\datanumber}:=\sqrt{\frac{1}{\datanumber}\sum_{i=1}^{\datanumber}\rbrtb{\functionnet_{\parameterv}(\inputvector_i)}^2}.
\end{equation*}

The Lipschitz property of neural networks is then as follows.
\begin{proposition}[Lipschitz property of neural networks]\label{Lipt2}
Assume that the  \emph{activation functions}~$\activefunctionb^l\,:\,\R^{\neuron_{l}}\to\R^{\neuron_{l}}$
are~$\actlipc$-Lipschitz with respect to the Euclidean norms on their input and output spaces. Then, it holds for every~$\inputvector\in\R^d$ and~$\parameterv=(\wtheta^\nhiddenlayer,\dots,\wtheta^0), \Gamma=(\wgamma^{L},\dots,\wgamma^{0}) \in\paraspace$ that
\begin{equation*}
   |{\functionnet_{\parameterv}(\inputvector)-\functionnet_{{\Gamma}}(\inputvector)}|\leq \Lipfunc \normFP{\parameterv-\Gamma}
\end{equation*}
with~$\Lipfunc:=2(\actlipc)^{L}\sqrt{\nhiddenlayer}\norm{\inputvector}\max_{l\in\cbrt{0,\dots,\nhiddenlayer}}\prod_{j\in\{0,\dots,\nhiddenlayer\},j\neq l}\rbrt{\norm{\weightmatrix^j}\vee\norm{\wgamma^j}}$. 

And similarly, it holds that
\begin{equation*}
    \normfcn{{\functionnet_{\parameterv}-\functionnet_{{\Gamma}}}}_{\datanumber}\leq \LipConst\normFP{\parameterv-\Gamma}
\end{equation*}
with~$\LipConst:=2(\actlipc)^{L}\sqrt{\nhiddenlayer}\normfcn{\inputvector}_{\datanumber}\max_{l\in\cbrt{0,\dots,\nhiddenlayer}}\prod_{j\in\{0,\dots,\nhiddenlayer\},j\neq l}\rbrt{\norm{\weightmatrix^j}\vee\norm{\wgamma^j}}$.
\end{proposition}
\noindent This property is helpful in bounding the quantiles of the empirical processes.
In particular, it can be used to show  that the networks are Lipschitz and bounded over typical sets that originate from our  regularization scheme.
Such a result is given in the following lemma.
\begin{lemma}[Lipschitz and boundedness on~$\paraspaceN$]\label{UPfunA1}
Under the conditions of Proposition~\ref{Lipt2},
it holds for  every~$\thetasym,\Gamma\in\paraspaceN$ that
\begin{equation*}
 \normfcn{\functionnet_\thetasym-\functionnet_\Gamma}_{\datanumber}\leq \LipConstone \normFP{\thetasym-\Gamma}
\end{equation*}
and that 
\begin{equation*}
 \normfcn{\functionnet_\thetasym}_{\datanumber}\leq \LipConstone 
\end{equation*}
with~$\LipConstone:=2(2\actlipc/L)^\nhiddenlayer\sqrt{\nhiddenlayer}\normfcn{\inputvector}_{\datanumber}$. 
\end{lemma}

To derive the complexity properties,
we denote  covering numbers by
$\covernpung(\radiusu,\mathcal{T},\mbox{\normfcn{\cdot}})$ and entropy by~$\entropy(\radiusu,\mathcal{T},\normfcn{\cdot}):=\log \covernpung(\radiusu,\mathcal{T},\normfcn{\cdot})$,
where~$\radiusu\in(0,\infty)$,~$\mathcal{T}$~is a set,
and~$\normfcn{\cdot}$~is a \mbox{(pseudo-)}norm on an ambient space of~$\mathcal{T}$~\citep[Page~98]{Vaart1996}. 
We use these numbers to define a complexity measure for a collection of networks~$\networksetgeneral:=\cbrt{\functionnet_\thetasym\,:\, \thetasym \in \subparaspace}$ by
\newcommand{\dudleyconst}{v}
\begin{equation}\label{entropyint}
    \entropyboundgeneral:= \int_{\radiusdelta/(8\sigma)}^{\infty}{\entropy^{1/2}\bigl(\radiusu,\networksetgeneral,\normfcn{\cdot}_\datanumber\bigr)d\radiusu}
\end{equation}
for~$\radiusdelta,\sigma\in(0,\infty)$ \citep[Section~3.3]{Sara2000}.
Almost in line with standard terminology, we call this complexity measure the \emph{Dudley integral}~\citep[Section~8.1]{Vershynin2018}.
We can bound the complexity of the class of neural networks~$\networksetN:=\cbrt{\functionnet_\thetasym:\thetasym\in\paraspaceN}$ that have parameters in the constraint set~$\paraspaceN$ as follows:

\begin{proposition}[Complexity properties of neural networks]\label{EN}
Assume that the \emph{activation functions} $\activefunctionb^l\,:\,\R^{\neuron_{l}}\to\R^{\neuron_{l}}$
are~$\actlipc$-Lipschitz continuous with respect to the Euclidean norms on their input and output spaces. 
Then, it holds for every $r\in(0,\infty)$ and~$\radiusdelta,\sigma\in(0,\infty)$ that satisfy~$\radiusdelta\leq 8\sigma\LipConstone$ that
\begin{equation*}\label{ENT0}
  \entropy\rbrtb{\radiusu,\networksetN,\normfcn{\cdot}_{\datanumber}}\leq   \frac{6(\LipConstone)^2 }{\radiusu^2}{\log\rbrtbbb{\frac{e\allparameters\radiusu^2}{(\LipConstone)^2}\vee 2e }}
\end{equation*}
and 
\begin{equation*}
    \entropyboundN\le \frac{5\LipConstone}{2} \sqrt{\log{(e\allparameters\vee 2e })}\log\rbrtbb{\frac{8\sigma\LipConstone}{\radiusdelta}},
\end{equation*}
where we  recall that~$\LipConstone=2(2\actlipc/L)^\nhiddenlayer\sqrt{\nhiddenlayer}\normfcn{\inputvector}_{\datanumber}$.
\end{proposition}

\section{Additional materials and proofs}
We now state some auxiliary results and then prove our claims.

\label{sec:proofs}
\subsection{Additional materials}
We first provide three auxiliary results that we use in our proofs.
We start with a slightly adapted version of~\citet[Corollary~8.3]{Sara2000}:
\begin{lemma}[Suprema over Gaussian processes]\label{lem:SupGau}
Consider a set~$\paraspaceprime\subset\paraspace$ 
and a  constant~$\uppboundfunc\in[0,\infty)$ such  that~$\sup_{\parameterv\in {\paraspaceprime}}\normfcn{\functionnet_{\parameterv}}_\datanumber\leq\uppboundfunc$.
Assume that the noise random variables~$\noise_1,\dots,\noise_n$ are independent, centered, and uniformly sub-Gaussian as specified on Page~\pageref{sgnoise}.
Then, there is a constant~$\constA\in (0,\infty)$ that depends only on~$\noisedis$ and~$\subGparaone$ such that for all~$\radiusdelta,\sigma \in (0,\infty)$  that satisfy~$\radiusdelta< \sigma\uppboundfunc$ and 
\begin{equation*}
    \sqrt{\datanumber}\radiusdelta\geq   \constA\bigl(\entropyboundprime\vee\uppboundfunc\bigr),
    \end{equation*}
it holds that
\begin{equation*}
   \mathbb{P} \biggl(\biggl\{\sup_{\parameterv\in {\paraspaceprime}}\biggl|\frac{1}{\datanumber}\sum_{i=1}^{\datanumber}{\functionnet_{\parameterv}(\inputvector_i)\noise_i}\biggr|\geq \radiusdelta\biggr\} \,\cap\, \biggl\{\frac{1}{\datanumber}\sum_{i=1}^{\datanumber}{{(\noise_i)}^2}\leq \sigma^2
   \biggr\}\biggr)\leq \constA e^{ -\frac{\datanumber\radiusdelta^2}{(\constA\uppboundfunc)^2}}.
\end{equation*}
\end{lemma}
This result is used to bound~$\tuning_{\ell_1,t}$.

We then turn to a Lipschitz property of metric entropy:
\begin{lemma}[Entropy transformation for Lipschitz functions]\label{generalLP}
Consider sets 
$\paraspaceprime\subset\paraspace$ and~$\networksetprime:=\{\functionnet_{\parameterv}\,:\,\parameterv\in\paraspaceprime\}$ and a  metric~$\parammetric\,:\,\paraspaceprime\times\paraspaceprime\to\R$. 
Assume that~$|\functionnet_{\parameterv}(\inputvector)-\functionnet_{\Gamma}(\inputvector)|\leq \Lipfcng(\inputvector) \parammetric(\parameterv,\Gamma)$ for  every~$\parameterv,\Gamma\in \paraspaceprime$ and~$\inputvector\in\R^{\inputdimention}$ and a  fixed function~$\Lipfcng\,:\,\R^\inputdimention\to[0,\infty)$.
Then,
\begin{equation*}
    \entropy(\radiusu,\networksetprime,\normfcn{\cdot}_{\datanumber})\leq \entropy\biggl(\frac{\radiusu}{\normfcn{\Lipfcng}_{\datanumber}},\paraspaceprime,{\parammetric}\biggr)~~~~~~\text{for all~}\radiusu\in(0,\infty),
\end{equation*}
where~$\normfcn{\Lipfcng}_{\datanumber}:=\sqrt{\sum_{i=1}^{\datanumber}(\Lipfcng(\inputvector_i))^2/\datanumber}$.
\end{lemma}
\noindent 
We use the  convention~$a/0=\infty$ for~$a\in(0,\infty)$. The result allows us to bound entropies on the parameter spaces instead of the network spaces.
We prove the lemma in the following section.

\newcommand{\subgausslevel}{v}
We conclude with a deviation inequality for the noise.
\begin{lemma}[Deviation of sub-Gaussian noise]\label{subgausslemma}
Assume that the noise variables $\noise_1,\dots,\allowbreak\noise_n$ are independent, centered, and uniformly sub-Gaussian as stipulated on Page~\pageref{subgauss}.
Then,
\begin{equation*}
  \mathbb{P}\biggl(\frac{1}{\datanumber}\sum_{i=1}^{\datanumber}{{(\noise_i})^2}\geq\subgausslevel\biggr)\le e^{-\frac{\datanumber \subgausslevel}{12\noisedis^2}}~~~~~~~\text{for all}~\subgausslevel\in[2\subGparaone^2,\infty).
\end{equation*}
\end{lemma}
\noindent This deviation inequality is tailored to our needs in the proof of Theorem~\ref{UPBOUA1}.

\subsection{Proofs}\label{proofs}
We provide here the proofs of our claims.
\subsubsection{Proof of Proposition~\ref{prop:equNNs}}
\begin{proof}
We prove the two directions in order.

\emph{Direction~1:}
Fix a~$\parameterv=(\weightmatrix^\nhiddenlayer,\dots,\weightmatrix^0)\in\paraspace$. 
Assume first   that~$\weightmatrix^l=\boldsymbol{0}_{\neuron_{l+1}\times\neuron_{l}}$ for an~$l\in\cbrt{0,\dots,\nhiddenlayer}$.
In view of the definition of neural networks in~\eqref{nns} and the assumed nonnegative homogeneity of the activation functions, it then holds that
\begin{multline*}
    \function_{\parameterv}(\inputvector)=\weightmatrix^\nhiddenlayer\activefunctionb^\nhiddenlayer\rbrt{\dots\boldsymbol{0}_{\neuron_{l+1}\times\neuron_{l}}\activefunctionb^l\rbrt{\dots\weightmatrix^1\activefunctionb^1\rbrt{\weightmatrix^0\inputvector}}}\\
    =\weightmatrix^\nhiddenlayer\activefunctionb^\nhiddenlayer\rbrt{\dots0\cdot\boldsymbol{0}_{\neuron_{l+1}\times\neuron_{l}}\activefunctionb^l\rbrt{\dots\weightmatrix^1\activefunctionb^1\rbrt{\weightmatrix^0\inputvector}}}=0
\end{multline*}
for all~$\inputvector\in\R^\inputdimention$.
Therefore,~$\kapasym:=0$ and all~$\thetasym\in\subparaspace$ satisfy~$\functionspace=\function_{\parameterv}$, as desired .

Assume now that~$\weightmatrix^l\neq\boldsymbol{0}_{\neuron_{l+1}\times\neuron_{l}}$ for all~$l\in\cbrt{0,\dots,\nhiddenlayer}$. 
Define~$\kapasym:=\rbrt{\generalfcn(\parameterv)}^{(\nhiddenlayer+1)/\homoorder}$
and~$\thetasym:=\parameterv/\kapasym^{1/(\nhiddenlayer+1)}=\rbrt{\weightmatrix^\nhiddenlayer/\kapasym^{1/(\nhiddenlayer+1)},\ldots,\weightmatrix^0/\kapasym^{1/(\nhiddenlayer+1)}}$ if~$\kapasym^{1/(\nhiddenlayer+1)}\neq 0$.
We need to show that~1.~$\kappa\in(0,\infty)$ and~$\thetasym\in\subparaspace$ and~2.~$\function_{\parameterv}=\functionspace$. 

Since~$\generalfcn$ is assumed positive definite,
it holds that~$\generalfcn(\parameterv)\in(0,\infty)$ and, therefore,~$\kapasym\in(0,\infty)$.
The fact that~$\kapasym>0$ also ensures that the parameter~$\thetasym$ is well-defined, and we can invoke the assumed nonnegative homogeneity of degree~$\homoorder$ of~$\generalfcn$
to derive
\begin{equation*}
    \generalfcn(\thetasym)=\generalfcn\rbrtb{\parameterv/\kapasym^{1/(\nhiddenlayer+1)}}=\rbrtb{\kapasym^{-1/(\nhiddenlayer+1)}}^\homoorder\generalfcn(\parameterv)=\rbrtb{\rbrt{\generalfcn(\parameterv)}^{(\nhiddenlayer+1)/\homoorder}}^{-\homoorder/(\nhiddenlayer+1)}\generalfcn(\parameterv)=1.
\end{equation*}
This verifies 1.

We can then invoke the assumed nonnegative homogeneity of degree~1 of the activation functions to derive for all~$\inputvector\in\R^\inputdimention$ that
\begin{align*}
    \functionspace(\inputvector)
    &=\kapasym\frac{\weightmatrix^\nhiddenlayer}{\kapasym^{1/(\nhiddenlayer+1)}}\activefunctionb^\nhiddenlayer\rbrtbbb{\dots\frac{\weightmatrix^1}{\kapasym^{1/(\nhiddenlayer+1)}}\activefunctionb^1\rbrtbb{\frac{\weightmatrix^0}{\kapasym^{1/(\nhiddenlayer+1)}}\inputvector}}\\
    &=\kapasym\frac{\weightmatrix^\nhiddenlayer}{\kapasym^{1/(\nhiddenlayer+1)}}\activefunctionb^\nhiddenlayer\rbrtbbb{\dots\frac{\weightmatrix^1}{\rbrtb{\kapasym^{1/(\nhiddenlayer+1)}}^2}\activefunctionb^1\rbrt{\weightmatrix^0\inputvector}}\\
    &=\dots\\
    &=\frac{\kapasym}{\rbrtb{\kapasym^{1/(\nhiddenlayer+1)}}^{(\nhiddenlayer+1)}}\weightmatrix^\nhiddenlayer\activefunctionb^\nhiddenlayer\rbrtb{\dots\weightmatrix^1\activefunctionb^1\rbrt{\weightmatrix^0\inputvector}}\\
    &=\weightmatrix^\nhiddenlayer\activefunctionb^\nhiddenlayer\rbrtb{\dots\weightmatrix^1\activefunctionb^1\rbrt{\weightmatrix^0\inputvector}}\\
    &=\function_{\parameterv}(\inputvector).
\end{align*}
This verifies 2.

\emph{Direction~2:} Fix a~$\kapasym\in[0,\infty)$ and a~$\thetasym=(\wmatrixsym^\nhiddenlayer,\dots,\wmatrixsym^0)\in\subparaspace$, 
and
define~$\parameterv:=\kapasym^{1/(\nhiddenlayer+1)}\thetasym=(\kapasym^{1/(\nhiddenlayer+1)}\wmatrixsym^\nhiddenlayer,\dots,\kapasym^{1/(\nhiddenlayer+1)}\wmatrixsym^0)$.
We then invoke the assumed nonnegative homogeneity of degree~1 of the activation functions to derive for all~$\inputvector\in\R^\inputdimention$ that
\begin{align*}
    \function_{\parameterv}(\inputvector)
    &=\kapasym^{1/(\nhiddenlayer+1)}\wmatrixsym^\nhiddenlayer\activefunctionb^\nhiddenlayer\rbrtbb{\dots\kapasym^{1/(\nhiddenlayer+1)}\wmatrixsym^1\activefunctionb^1\rbrtb{\kapasym^{1/(\nhiddenlayer+1)}\wmatrixsym^0\inputvector}}\\
    &=\kapasym^{1/(\nhiddenlayer+1)}\wmatrixsym^\nhiddenlayer\activefunctionb^\nhiddenlayer\rbrtbb{\dots\rbrtb{\kapasym^{1/(\nhiddenlayer+1)}}^2\wmatrixsym^1\activefunctionb^1\rbrt{\wmatrixsym^0\inputvector}}\\
    &=\dots\\
    &=\rbrtb{\kapasym^{1/(\nhiddenlayer+1)}}^{(\nhiddenlayer+1)}\wmatrixsym^\nhiddenlayer\activefunctionb^\nhiddenlayer\rbrtb{\dots\wmatrixsym^1\activefunctionb^1\rbrt{\wmatrixsym^0\inputvector}}\\
    &=\kapasym\wmatrixsym^\nhiddenlayer\activefunctionb^\nhiddenlayer\rbrtb{\dots\wmatrixsym^1\activefunctionb^1\rbrt{\wmatrixsym^0\inputvector}}\\
    &=\functionspace(\inputvector),
\end{align*}
as desired.
\end{proof}

\subsubsection{Proof of Theorem~\ref{GPbound}}
\begin{proof}
Since~$(\estimatedkappag,\estimatedthetag)$ is a minimizer of the objective function in~\eqref{estimator}, 
we find for every~$\kapasym\in[0,\infty)$ and~$\thetasym\in\subparaspace$ that
\begin{equation*}
    \frac{1}{\datanumber}\sum_{i=1}^{\datanumber}\bigl(\outputy_i-\estimatedfunctionnew(\inputvector_i)\bigr)^2+\tuning\estimatedkappag\leq \frac{1}{\datanumber}\sum_{i=1}^{\datanumber}\bigl(\outputy_i-\kapasym \functionnet_{\thetasym}(\inputvector_i)\bigr)^2+\tuning\kapasym.
\end{equation*}
Replacing the~$\outputy_i$'s via the model in~\eqref{model} then yields
\begin{equation*}
  \frac{1}{\datanumber}\sum_{i=1}^{\datanumber}\bigl(\functiontruei+\noise_i-\estimatedfunctionnew(\inputvector_i)\bigr)^2+\tuning\estimatedkappag\leq \frac{1}{\datanumber}\sum_{i=1}^{\datanumber}\bigl(\functiontruei+\noise_i-\kapasym \functionnet_{\thetasym}(\inputvector_i)\bigr)^2+\tuning\kapasym.
\end{equation*}
Expanding the squared-terms and rearranging terms, we get
\begin{multline*}
   \frac{1}{\datanumber}\sum_{i=1}^{\datanumber}\bigl(\estimatedfunctionnew(\inputvector_i)-\functiontruei\bigr)^2\leq \frac{1}{\datanumber}\sum_{i=1}^{\datanumber}\bigl(\kapasym \functionnet_{\thetasym}(\inputvector_i)-\functiontruei\bigr)^2\\
   +\frac{2}{\datanumber}\sum_{i=1}^{\datanumber}{\estimatedfunctionnew(\inputvector_i) \noise_i}-\frac{2}{\datanumber}\sum_{i=1}^{\datanumber}{\kapasym \functionnet_{\thetasym}(\inputvector_i) \noise_i}+\tuning\kapasym-\tuning\estimatedkappag.  
\end{multline*}
We can then bound the sums on the second line to obtain
\begin{multline*}
   \frac{1}{\datanumber}\sum_{i=1}^{\datanumber}\bigl(\estimatedfunctionnew(\inputvector_i)-\functiontruei\bigr)^2\leq \frac{1}{\datanumber}\sum_{i=1}^{\datanumber}\bigl(\kapasym \functionnet_{\thetasym}(\inputvector_i)-\functiontruei\bigr)^2\\
   +\estimatedkappag\sup_{\parasup\in\subparaspace}\Bigl|\frac{2}{\datanumber}\sum_{i=1}^{\datanumber}{\functionnet_{\parasup}(\inputvector_i) \noise_i}\Bigr|+\kapasym\sup_{\parasup\in\subparaspace}\Bigl|\frac{2}{\datanumber}\sum_{i=1}^{\datanumber}{\functionnet_{\parasup}(\inputvector_i) \noise_i}\Bigr|+\tuning\kapasym-\tuning\estimatedkappag.  
\end{multline*}
The second line can then be consolidated by virtue of the assumption on~$\tuning$:
with probability at least~$1-\level$,
it holds that
\begin{equation*}
   \frac{1}{\datanumber}\sum_{i=1}^{\datanumber}\bigl(\estimatedfunctionnew(\inputvector_i)-\functiontruei\bigr)^2\leq \frac{1}{\datanumber}\sum_{i=1}^{\datanumber}\bigl(\kapasym \functionnet_{\thetasym}(\inputvector_i)-\functiontruei\bigr)^2
   +2\tuning\kapasym.
\end{equation*}
Taking the infimum over~$\kapasym\in[0,\infty)$ and~$\thetasym\in\subparaspace$ and invoking the definition of~$\prederrors{\cdot}$ on Page~$\pageref{eq:prederrors}$ gives the desired result.
\end{proof}
\subsubsection{Proof of Theorem~\ref{UPBOUA1}}
\begin{proof}
The idea of the proof is to bound the effective noise and then apply  Theorem~\ref{GPbound}.

If~$\LipConstone=0$, then~$\functionnet_{\thetasym}(\inputvector_i)=0$ for all~$\thetasym\in\paraspaceN$ and~$i\in\{1,\dots,\datanumber\}$  in view of Lemma~\ref{UPfunA1}. 
Hence,
\begin{equation*}
   \mathbb{P} \biggl(\sup_{\thetasym\in {\paraspaceN}}\biggl|\frac{2}{\datanumber}\sum_{i=1}^{\datanumber}{\functionnet_{\thetasym}(\inputvector_i)\noise_i}\biggr|\leq  \radiusdelta\biggr)=\mathbb{P}(0\leq  \radiusdelta)=1~~~~~\text{for all~}\radiusdelta\in(0,\infty),
\end{equation*}
which makes a proof straightforward.
We can thus assume~$\LipConstone>0$ in the following.
 
Our first step is to apply Lemma~\ref{lem:SupGau} about suprema of empirical processes with~$\paraspaceprime:=\paraspaceN$.
For this, we need to find 1.~a constant~$\uppboundfunc\in[0,\infty)$ that satisfies~$\sup_{\thetasym\in\paraspaceN}\normfcn{\functionnet_{\thetasym}}_\datanumber\leq\uppboundfunc$ and 
2.~suitable~$\radiusdelta,\sigma\in(0,\infty)$ that satisfy~$\radiusdelta<\sigma\uppboundfunc$ and  
\begin{equation*}
     \sqrt{\datanumber}\geq \frac{\constA}{\radiusdelta} \bigl( \entropyboundN\vee\uppboundfunc\bigr).
\end{equation*}
 Condition~1 is verified by~$\uppboundfunc:=\LipConstone$ according to Lemma~\ref{UPfunA1}.

For Condition~2, we define~$\radiusdelta\equiv\radiusdelta(\datanumber,\allparameters,\constA,\LipConstone):=10\constA\LipConstone  \sqrt{\log(2\allparameters)/\datanumber}\log(2\datanumber)\in(0,\infty)$ and~$\sigma:=(2\radiusdelta/\LipConstone)\vee (\sqrt{2}\subGparaone)$.
Then,
$\radiusdelta<\sigma\uppboundfunc$ 
by the definitions of~$\radiusdelta,\sigma,\uppboundfunc$.
Moreover,
Proposition~\ref{EN}, the definitions  of~$\radiusdelta$ and~$\uppboundfunc$,
and
\begin{equation*}
    \frac{8\sigma\LipConstone}{\delta}= 16\vee\frac{8\sqrt{2}\subGparaone\LipConstone}{10\constA\LipConstone\sqrt{\log(2\allparameters)/\datanumber}\log(2\datanumber)}\leq 16\vee\frac{2\subGparaone\sqrt{\datanumber}}{\constA}\leq16\sqrt{\datanumber}
\end{equation*}
(we  assume that~$\constA\geq \gamma/8$ without loss of generality and use that~$\log(2)\geq 0.69$)
yield
\begin{align*}
    &\frac{\constA}{\radiusdelta}\bigl(\entropyboundN\vee\LipConstone\bigr)\\
 &\leq\frac{\constA}{\radiusdelta}\biggl(\frac{5\LipConstone}{2}\sqrt{\log\rbrtb{e\allparameters\vee 2e}}\log\rbrtbb{\frac{8\sigma\LipConstone}{\radiusdelta}}\vee \LipConstone\biggr)\\
 &\leq\frac{\sqrt{\datanumber}\constA}{10\constA\LipConstone  \sqrt{\log(2\allparameters)}\log(2\datanumber)}
 \Bigl(\frac{5\LipConstone}{2}\sqrt{\log(e\allparameters\vee 2e)}
 \log(16\sqrt\datanumber)\vee \LipConstone\Bigr)\\
 &\le \frac{\sqrt{\datanumber}\constA}{10\constA\LipConstone  \sqrt{\log(2\allparameters)}\log(2\datanumber)}
 \bigl(10\LipConstone\sqrt{\log(2\allparameters)}\log(2\datanumber)\vee \LipConstone\bigr)\\
 &=\sqrt{\datanumber},
\end{align*}
which verifies Condition~2.

We can thus apply Lemma~\ref{lem:SupGau} with the above-specified parameters to obtain that
\begin{equation*}
   \mathbb{P} \biggl(\biggl\{\sup_{\thetasym\in {\paraspaceN}}\biggl|\frac{1}{\datanumber}\sum_{i=1}^{\datanumber}{\functionnet_{\thetasym}(\inputvector_i)\noise_i}\biggr|\geq \radiusdelta\biggr\} \,\cap\, \biggl\{\frac{1}{\datanumber}\sum_{i=1}^{\datanumber}{(\noise_i)^2}\leq \sigma^2
   \biggr\}\biggr)\leq \constA e^{-\frac{\datanumber\radiusdelta^2}{(\constA\LipConstone)^2} }.
\end{equation*}
We use  that~$\mathbb{P}(\mathcal C \cap \mathcal D)\le \alpha$ implies~$\mathbb{P}(\mathcal C^\complement \cup \mathcal D^\complement)\ge 1-\alpha$ and that~$\mathbb{P}(\mathcal C^\complement)\ge \mathbb{P}(\mathcal C^\complement \cup \mathcal D^\complement)-\mathbb{P}(\mathcal D^\complement)$ to rewrite this inequality as
\begin{multline*}
\mathbb{P} \biggl(\sup_{\thetasym\in {\paraspaceN}}\biggl|\frac{1}{\datanumber}\sum_{i=1}^{\datanumber}{\functionnet_{\thetasym}(\inputvector_i)\noise_i}\biggr|\leq  \radiusdelta \biggr)
\ge
   \mathbb{P} \biggl(\sup_{\thetasym\in {\paraspaceN}}\biggl|\frac{1}{\datanumber}\sum_{i=1}^{\datanumber}{\functionnet_{\thetasym}(\inputvector_i)\noise_i}\biggr|< \radiusdelta \biggr)\\
 \ge 1-\constA e^{ -\frac{\datanumber\radiusdelta^2}{(\constA\LipConstone)^2}}- 
  \mathbb{P} \biggl(\frac{1}{\datanumber}\sum_{i=1}^{\datanumber}{(\noise_i)^2}> \sigma^2\biggr).
\end{multline*}
Since~$\sigma^2\geq 2\gamma^2$ by the definition of~$\sigma$, Lemma~\ref{subgausslemma} with~$\subgausslevel:=\sigma^2$ allows us to bound the last term according  to
\begin{equation*}
  \mathbb{P}\biggl(\frac{1}{\datanumber}\sum_{i=1}^{\datanumber}{(\noise_i)^2}>\sigma^2\biggr)\leq  \mathbb{P}\biggl(\frac{1}{\datanumber}\sum_{i=1}^{\datanumber}{(\noise_i)^2}\geq \sigma^2\biggr) \le e^{-\frac{\datanumber 
  \sigma^2}{12\noisedis^2}}.
\end{equation*}
Combining this inequality with the previous one yields
\begin{equation*}
   \mathbb{P} \biggl(\sup_{\thetasym\in {\paraspaceN}}\biggl|\frac{1}{\datanumber}\sum_{i=1}^{\datanumber}{\functionnet_{\thetasym}(\inputvector_i)\noise_i}\biggr|\leq  \radiusdelta \biggr)
  \ge 1-\constA e^{ -\frac{\datanumber\radiusdelta^2}{(\constA\LipConstone)^2}}- e^{-\frac{\datanumber 
   \sigma^2}{12\noisedis^2}}.
 \end{equation*}
 
By the definitions of~$\radiusdelta$ and~$\sigma$,
 and assuming that~$\datanumber$ is large enough (depending on~$\subGparaone,\noisedis$),
we find that
\begin{align*}
  &\constA e^{ -\frac{\datanumber\radiusdelta^2}{(\constA\LipConstone)^2}}+ e^{-\frac{\datanumber 
   \sigma^2}{12\noisedis^2}}\\
   &\leq \constA e^{ -10^2\log(2\allparameters)(\log(2\datanumber))^2}+
   e^{-\frac{4\cdot 10^2(\constA)^2\log(2\allparameters)(\log(2\datanumber))^2}{12\noisedis^2}}+
   e^{-\frac{\datanumber\subGparaone^2 
   }{6\noisedis^2}}\\
   &\leq e^{ -\log(4\datanumber)}+
   e^{-\log(4\datanumber)}+
   e^{-\frac{\datanumber\subGparaone^2 
   }{6\noisedis^2}}\\
   &\leq \frac{1}{\datanumber},
\end{align*}
that is,
 \begin{equation*}
   \mathbb{P} \biggl(\sup_{\thetasym\in {\paraspaceN}}\biggl|\frac{1}{\datanumber}\sum_{i=1}^{\datanumber}{\functionnet_{\thetasym}(\inputvector_i)\noise_i}\biggr|\leq  \radiusdelta \biggr)
  \ge 1-\frac{1}{\datanumber}.
 \end{equation*}
 In other words, $\tuningbound\leq2\radiusdelta$ for~$\level=1/\datanumber$.

 The claim then follows directly from Theorem~\ref{GPbound}  with~$\tuning\geq2\radiusdelta=20\constA\LipConstone \allowbreak \sqrt{\log(2\allparameters)/\datanumber}\allowbreak\log(2\datanumber)$,  $\LipConstone=2(2\actlipc/L)^\nhiddenlayer\sqrt{\nhiddenlayer}\normfcn{\inputvector}_{\datanumber}$ (see Lemma~\ref{UPfunA1}), and~$\constAtwo:=40\constA$.
\end{proof}
\subsubsection{Proof of Lemma~\ref{inSamVsOutSam}}
\begin{proof}
The idea of the proof is to disentangle the generalization error into the prediction error and additional terms.
We then bound the prediction error by using Corollary~\ref{Paramset}  and the additional terms by using empirical-process theory.
 
We  1.~replace the output~$\outputy$ by using the model in~\eqref{model}, 
2.~use monotone  convergence together with the fact that $(r+s)^2\le  \MCN r^2+1.01s^2$ for a numerical constant $b\in(0,\infty)$, 
3.~use the linearity of expectations, 
4.~add a zero-valued term, 
and 5.~take an absolute value  to get
\begingroup
\allowdisplaybreaks
\begin{align*}
&\E_{(\inputvector,\outputy)}\Bigl[\bigl(\estimatedfunctionnew(\inputvector)-\outputy\bigr)^2\Bigr]\\
&=\E_{(\inputvector,\outputy)}\Bigl[\bigl(\estimatedfunctionnew(\inputvector)- \functiontrue(\inputvector)-\noise\bigr)^2\Bigr]\\
&\leq\E_{(\inputvector,\outputy)}\Bigl[\MCN\bigl(\estimatedfunctionnew(\inputvector)- \functiontrue(\inputvector)\bigr)^2+1.01\noise^2\Bigr]\\
&=\MCN\E_{(\inputvector,\outputy)}\Bigl[\bigl(\estimatedfunctionnew(\inputvector)- \functiontrue(\inputvector)\bigr)^2\Bigr]+1.01\E_{(\inputvector,\outputy)}[\noise^2]\\
&= \frac{\MCN}{\datanumber}\sum_{i=1}^{\datanumber}\bigl(\estimatedfunctionnew(\inputvector_i)-\functiontruei\bigr)^2+1.01\E_{(\inputvector,\outputy)}[\noise^2]\\
&~~~~~~~- \frac{\MCN}{\datanumber}\sum_{i=1}^{\datanumber}\bigl(\estimatedfunctionnew(\inputvector_i)-\functiontruei\bigr)^2+\MCN\E_{(\inputvector,\outputy)}\Bigl[\bigl(\estimatedfunctionnew(\inputvector)- \functiontrue(\inputvector)\bigr)^2\Bigr]\\
&\leq \frac{\MCN}{\datanumber}\sum_{i=1}^{\datanumber}\bigl(\estimatedfunctionnew(\inputvector_i)-\functiontruei\bigr)^2+1.01\E_{(\inputvector,\outputy)}[\noise^2]\\
&~~~~~~~+\bigg|\frac{\MCN}{\datanumber}\sum_{i=1}^{\datanumber}\bigl(\estimatedfunctionnew(\inputvector_i)-\functiontruei\bigr)^2-\MCN\E_{(\inputvector,\outputy)}\Bigl[\bigl(\estimatedfunctionnew(\inputvector)- \functiontrue(\inputvector)\bigr)^2\Bigr]\biggr |.
\end{align*}
\endgroup

The remaining challenge is now to  bound the last term of this display.
We devise an approach based on symmetrization for probabilities~\citep[Lemma~16.1]{van2016estimation}.  
We first use 1.~the fact that $\estimatedthetag\in\paraspaceN$ and
2.~the independence assumption on the data to get
\begingroup
\allowdisplaybreaks
\begin{align*}
 &\biggl|\frac{\MCN}{\datanumber}\sum_{i=1}^{\datanumber}\bigl(\estimatedfunctionnew(\inputvector_i)-\functiontruei\bigr)^2-\MCN\E_{(\inputvector,\outputy)}\Bigl[\bigl(\estimatedfunctionnew(\inputvector)- \functiontrue(\inputvector)\bigr)^2\Bigr]\biggr|\\
  &\le\sup_{\thetasym\in {\paraspaceN}}\biggl|\frac{\MCN}{\datanumber}\sum_{i=1}^{\datanumber}\bigl(\estimatedkappag\functionnet_{\parasup}(\inputvector_i)-\functiontruei\bigr)^2-\MCN\E_{(\inputvector,\outputy)}\Bigl[\bigl(\estimatedkappag\functionnet_{\parasup}(\inputvector)- \functiontrue(\inputvector)\bigr)^2\Bigr]\biggr|\\
 &=\sup_{\thetasym\in {\paraspaceN}}\biggl|\frac{\MCN}{\datanumber}\sum_{i=1}^{\datanumber}\Bigl(\bigl(\estimatedkappag\functionnet_{\parasup}(\inputvector_i)-\functiontruei\bigr)^2-\E_{(\inputvector_1,\outputy_1),\dots,(\inputvector_\datanumber,\outputy_\datanumber)}\Bigl[\bigl(\estimatedkappag\functionnet_{\parasup}(\inputvector_i)- \functiontrue(\inputvector_i)\bigr)^2\Bigr]\Bigr)\biggr|.
\end{align*}
\endgroup

We now prepare the application of ~\citet[Lemma~16.1]{van2016estimation}.
We use
1.~the definition of $(\SyProP)^2$, which is called ``$R^2$’'  in~\citet[Lemma~16.1]{van2016estimation}, 
2.~the fact that $(r+s)^4\le 8r^4+8s^4$ and dominated convergence,
 3.~the fact that $\kapastarone\functionnet_{\thetastarone}(\inputvector_i)=\functiontrue(\inputvector_i)$ by assumption and the linearity of finite sums and expectations,
  4.~again the linearity of finite sums and expectations, the fact that $\thetastarone\in\paraspaceN$, and dominated convergence, 
  5.~the fact that $\sum_{i=1}^{\datanumber}(\functionnet_{\parasup}(\inputvector_i))^4/\datanumber\le 16(2\actlipc/L)^{4\nhiddenlayer}\nhiddenlayer^2 \sum_{i=1}^{\datanumber}\norm{\inputvector_i}^4/\datanumber$ and analogs of Proposition~\ref{Lipt2}  and Lemma~\ref{UPfunA1}, 
  6.~$\estimatedkappag\leq 3\kapastarone$, which can be proved easily along the lines of the proof of Theorem~\ref{GPbound} (just double the tuning parameter), 
and once more the linearity of integrals, and
7.~a simplification with  a numerical constant $\tempConst\in(0,\infty)$, which may change from line to line in the proof, to obtain
\begingroup
\allowdisplaybreaks
\begin{align*}
 (\SyProP)^2&=\sup_{\thetasym\in {\paraspaceN}}\frac{1}{\datanumber}\sum_{i=1}^{\datanumber}\E_{(\inputvector_i,\outputy_i)}\Bigl[\bigl(\estimatedkappag\functionnet_{\parasup}(\inputvector_i)-\functiontruei\bigr)^4\Bigr]\\
 &\le\sup_{\thetasym\in {\paraspaceN}}\frac{1}{\datanumber}\sum_{i=1}^{\datanumber}\E_{(\inputvector_i,\outputy_i)}\Bigl[8\bigl(\estimatedkappag\functionnet_{\parasup}(\inputvector_i)\bigr)^4+8\bigl(\functiontruei\bigr)^4\Bigr]\\
 &=8(\estimatedkappag)^4\sup_{\thetasym\in {\paraspaceN}}\frac{1}{\datanumber}\sum_{i=1}^{\datanumber}\E_{(\inputvector_i,\outputy_i)}\Bigl[\bigl(\functionnet_{\parasup}(\inputvector_i)\bigr)^4\Bigr]+\frac{8(\kapastarone)^4}{\datanumber}\sum_{i=1}^{\datanumber}\E_{(\inputvector_i,\outputy_i)}\Bigl[\bigl(\functionnet_{\thetastarone}(\inputvector_i)\bigr)^4\Bigr]\\
 &\leq 8\bigl((\estimatedkappag)^4+(\kapastarone)^4\bigl)\E_{(\inputvector_1,\outputy_1),\dots,(\inputvector_\datanumber,\outputy_\datanumber)}\biggl[\sup_{\thetasym\in {\paraspaceN}}\frac{1}{\datanumber}\sum_{i=1}^{\datanumber}\bigl(\functionnet_{\parasup}(\inputvector_i)\bigr)^4\biggr]\\
            &\leq 8\bigl((\estimatedkappag)^4+(\kapastarone)^4\bigl)\E_{(\inputvector_1,\outputy_1),\dots,(\inputvector_\datanumber,\outputy_\datanumber)}\Biggl[16\biggl(\frac{2\actlipc}{\nhiddenlayer}\biggr)^{4\nhiddenlayer}\frac{\nhiddenlayer^2}{\datanumber}\sum_{i=1}^{\datanumber}\norm{\inputvector_i}^4\Biggr]\\
 &\le128\bigl((3\kapastarone)^4+(\kapastarone)^4\bigl)\biggl(\frac{2\actlipc}{\nhiddenlayer}\biggr)^{4\nhiddenlayer}\nhiddenlayer^2\E_{(\inputvector_1,\outputy_1),\dots,(\inputvector_\datanumber,\outputy_\datanumber)}\biggl[\frac{1}{\datanumber}\sum_{i=1}^{\datanumber}\norm{\inputvector_i}^4\biggr]\\
 &=\tempConst(\kapastarone)^4\biggl(\frac{2\actlipc}{\nhiddenlayer}\biggr)^{4\nhiddenlayer}\nhiddenlayer^2\E_{(\inputvector_1,\outputy_1),\dots,(\inputvector_\datanumber,\outputy_\datanumber)}\biggl[\frac{1}{\datanumber}\sum_{i=1}^{\datanumber}\norm{\inputvector_i}^4\biggr].
\end{align*}
\endgroup

Then, we use 1.~the penultimate inequality and a rearrangement,
2.~the symmetrization bound of~\citet[Lemma~16.1]{van2016estimation} 
with an i.i.d.~Rademacher variables $\rademacherZeta_1,\dots,\rademacherZeta_\datanumber\in\{\pm1\}$ that are independent of the data,
3.~multiplying by a one-valued factor, 
4~the contraction principle~\citep[Theorem~4.4]{Ledoux1991} with $\alpha_i=(\estimatedkappag\functionnet_{\parasup}(\inputvector_i)-\functiontruei)^2/(2(\estimatedkappag\functionnet_{\parasup}(\inputvector_i))^2+2(\functiontruei)^2)\in [0,1]$,
5.~the fact that $\mathbb{P}(r+s>w)\le \mathbb{P}(r>w/2)+\mathbb{P}(s>w/2)$, 
6.~the linearity of finite sums, ~$\kapastarone\functionnet_{\thetastarone}(\inputvector_i)=\functiontrue(\inputvector_i)$, and the fact that $\thetastarone\in\paraspaceN$, and 
7.~$\estimatedkappag\leq 3\kapastarone$ to get for all $\range\in[4,\infty)$ 
\begingroup
\allowdisplaybreaks
\begin{align*}
 &\mathbb{P}\Biggl(\biggl|\frac{\MCN}{\datanumber}\sum_{i=1}^{\datanumber}\bigl(\estimatedfunctionnew(\inputvector_i)-\functiontruei\bigr)^2-\MCN\E_{(\inputvector,\outputy)}\Bigl[\bigl(\estimatedfunctionnew(\inputvector)- \functiontrue(\inputvector)\bigr)^2\Bigr]\biggr|> 4\MCN\SyProP\sqrt{\frac{2\range}{\datanumber}} \Biggr)\\
 &\le\mathbb{P}\Biggl(\sup_{\thetasym\in {\paraspaceN}}\biggl|\frac{1}{\datanumber}\sum_{i=1}^{\datanumber}\Bigl(\bigl(\estimatedkappag\functionnet_{\parasup}(\inputvector_i)-\functiontruei\bigr)^2-\E_{(\inputvector_1,\outputy_1),\dots,(\inputvector_\datanumber,\outputy_\datanumber)}\Bigl[\bigl(\estimatedkappag\functionnet_{\parasup}(\inputvector_i)- \functiontrue(\inputvector_i)\bigr)^2\Bigr]\Bigr)\biggr|> 4\SyProP\sqrt{\frac{2\range}{\datanumber}}\Biggr)\\
  &\le4\mathbb{P}\Biggl(\sup_{\thetasym\in {\paraspaceN}}\biggl|\frac{1}{\datanumber}\sum_{i=1}^{\datanumber}\rademacherZeta_i\bigl(\estimatedkappag\functionnet_{\parasup}(\inputvector_i)-\functiontruei\bigr)^2\biggr|>  \SyProP\sqrt{\frac{2\range}{\datanumber}}\Biggr)\\
   &=4\mathbb{P}\Biggl(\sup_{\thetasym\in {\paraspaceN}}\Biggl|\frac{1}{\datanumber}\sum_{i=1}^{\datanumber}\rademacherZeta_i\Bigl(2\bigl(\estimatedkappag\functionnet_{\parasup}(\inputvector_i)\bigr)^2+2\bigl(\functiontruei\bigr)^2\Bigr)\frac{\bigl(\estimatedkappag\functionnet_{\parasup}(\inputvector_i)-\functiontruei\bigr)^2}{2\bigl(\estimatedkappag\functionnet_{\parasup}(\inputvector_i)\bigr)^2+2\bigl(\functiontruei\bigr)^2}\Biggr|>  \SyProP\sqrt{\frac{2\range}{\datanumber}}\Biggr)\\
    &\le8\mathbb{P}\Biggl(\sup_{\thetasym\in {\paraspaceN}}\biggl|\frac{1}{\datanumber}\sum_{i=1}^{\datanumber}\rademacherZeta_i\Bigl(2\bigl(\estimatedkappag\functionnet_{\parasup}(\inputvector_i)\bigr)^2+2\bigl(\functiontruei\bigr)^2\Bigr)\biggr|>  \SyProP\sqrt{\frac{2\range}{\datanumber}}\Biggr)\\
   &\le8\mathbb{P}\Biggl(\sup_{\thetasym\in {\paraspaceN}}\biggl|\frac{1}{\datanumber}\sum_{i=1}^{\datanumber}2\rademacherZeta_i \bigl(\estimatedkappag\functionnet_{\parasup}(\inputvector_i)\bigr)^2\biggr|> \frac{\SyProP}{2}\sqrt{\frac{2\range}{\datanumber}}\Biggr)+8\mathbb{P}\Biggl(\biggl|\frac{1}{\datanumber}\sum_{i=1}^{\datanumber}2\rademacherZeta_i \bigl(\functiontrue(\inputvector_i)\bigr)^2\biggr|>\frac{\SyProP}{2}\sqrt{\frac{2\range}{\datanumber}}\Biggr)\\
    &=8\mathbb{P}\Biggl(\sup_{\thetasym\in {\paraspaceN}}\biggl|\frac{2(\estimatedkappag)^2}{\datanumber}\sum_{i=1}^{\datanumber}\rademacherZeta_i\bigl(\functionnet_{\parasup}(\inputvector_i)\bigr)^2\biggr|>\frac{\SyProP}{2}\sqrt{\frac{2\range}{\datanumber}}\Biggr)+8 \mathbb{P}\Biggl(\sup_{\thetasym\in {\paraspaceN}}\biggl|\frac{2(\kapastarone)^2}{\datanumber}\sum_{i=1}^{\datanumber}\rademacherZeta_i\bigl(\functionnet_{\parasup}(\inputvector_i)\bigr)^2\biggr|> \frac{\SyProP}{2}\sqrt{\frac{2\range}{\datanumber}}\Biggr)\\
    &\le16 \mathbb{P}\Biggl(\sup_{\thetasym\in {\paraspaceN}}\biggl|\frac{18(\kapastarone)^2}{\datanumber}\sum_{i=1}^{\datanumber}\rademacherZeta_i\bigl(\functionnet_{\parasup}(\inputvector_i)\bigr)^2\biggr|> \frac{\SyProP}{2}\sqrt{\frac{2\range}{\datanumber}}\Biggr).
\end{align*}
\endgroup
In the case $\kapastarone=0$, 
the probability equals zero (notice that $\SyProP\in[0,\infty)$, $\range\in[4,\infty)$, and $\datanumber\in[1,\infty)$), which is commensurate with the bound in Lemma~\ref{inSamVsOutSam}.
So, for the rest of the proof we can assume without loss of generality  that $\kapastarone> 0$.
Rearranging the above display then gives 
\begin{multline*}
 \mathbb{P}\Biggl(\biggl|\frac{\MCN}{\datanumber}\sum_{i=1}^{\datanumber}\bigl(\estimatedfunctionnew(\inputvector_i)-\functiontruei\bigr)^2-\MCN\E_{(\inputvector,\outputy)}\Bigl[\bigl(\estimatedfunctionnew(\inputvector)- \functiontrue(\inputvector)\bigr)^2\Bigr]\biggr|> 4\MCN\SyProP\sqrt{\frac{2\range}{\datanumber}} \Biggr)\\   
  \le16 \mathbb{P}\Biggl(\sup_{\thetasym\in {\paraspaceN}}\biggl|\frac{1}{\datanumber}\sum_{i=1}^{\datanumber}\rademacherZeta_i\bigl(\functionnet_{\parasup}(\inputvector_i)\bigr)^2\biggr|> \frac{\SyProP}{36(\kapastarone)^2}\sqrt{\frac{2\range}{\datanumber}}\Biggr).
\end{multline*}

Following the same approach as in the proof of Theorem~\ref{UPBOUA1} (with $\radiusdelta=1280\constA (2\actlipc/L)^{2\nhiddenlayer}\allowbreak \sqrt{\nhiddenlayer ^2\log(2\allparameters)\sum_{i=1}^{\datanumber}\norm{\inputvector_i}^4/\datanumber^2}\log(2\datanumber) $), we get 
\begin{equation*}
    \mathbb{P}\Biggl(\sup_{\thetasym\in {\paraspaceN}}\biggl|\frac{1}{\datanumber}\sum_{i=1}^{\datanumber}\rademacherZeta_i\bigl(\functionnet_{\parasup}(\inputvector_i)\bigr)^2\biggr|\leq\frac{\SyProP}{36(\kapastarone)^2}\sqrt{\frac{2\range}{\datanumber}}\Biggr)\geq1-\frac{1}{32\datanumber}
\end{equation*}
for $\range:=(\tempConst\constA(\kapastarone)^2(2\actlipc/L)^{2\nhiddenlayer} \sqrt{\nhiddenlayer ^2\log(2\allparameters)\sum_{i=1}^{\datanumber}\norm{\inputvector_i}^4/\datanumber}\log(2\datanumber)/\SyProP)^2/2$.
(note that  $\range\in[4,\infty)$ as long as $\datanumber$ is large enough such that $\tempConst(\constA)^2\log(2\allparameters)(\log(2\datanumber))^2\ge 4$; and also let remind our assumption in Theorem~\ref{UPBOUA1} that $\constA\geq \gamma/8$ ). 
Hence, we obtain
\begin{equation*}
    \mathbb{P}\Biggl(\sup_{\thetasym\in {\paraspaceN}}\biggl|\frac{1}{\datanumber}\sum_{i=1}^{\datanumber}\rademacherZeta_i\bigl(\functionnet_{\parasup}(\inputvector_i)\bigr)^2\biggr|>\frac{\SyProP}{36(\kapastarone)^2}\sqrt{\frac{2\range}{\datanumber}}\Biggr)< \frac{1}{32\datanumber}.
\end{equation*}
Now, we combine the above inequality with the previous result and using some rearrangements to obtain
\begin{equation*}
     \mathbb{P}\Biggl(\biggl|\frac{\MCN}{\datanumber}\sum_{i=1}^{\datanumber}\Bigl(\bigl(\estimatedfunctionnew(\inputvector_i)-\functiontruei\bigr)^2\Bigr)-\MCN\E_{(\inputvector,\outputy)}\Bigl[\bigl(\estimatedfunctionnew(\inputvector)- \functiontrue(\inputvector)\bigr)^2\Bigr]\biggr|> 4\MCN\SyProP\sqrt{\frac{2\range}{\datanumber}} \Biggr)< \frac{1}{2\datanumber}.
\end{equation*}
Collecting all pieces of the proof, we obtain 
\begingroup
\allowdisplaybreaks 
\begin{align*}
\E&_{(\inputvector,\outputy)}\Bigl[\bigl(\estimatedfunctionnew(\inputvector)-\outputy\bigr)^2\Bigr]\\
&\le \frac{\MCN}{\datanumber}\sum_{i=1}^{\datanumber}\bigl(\estimatedfunctionnew(\inputvector_i)-\functiontruei\bigr)^2+1.01\E_{(\inputvector,\outputy)}[\noise^2]\\&~~~~~~~+\biggl|\frac{\MCN}{\datanumber}\sum_{i=1}^{\datanumber}\bigl(\estimatedfunctionnew(\inputvector_i)-\functiontruei\bigr)^2-\MCN\E_{(\inputvector,\outputy)}\Bigl[\bigl(\estimatedfunctionnew(\inputvector)- \functiontrue(\inputvector)\bigr)^2\Bigr]\biggr|\\
&\le \frac{\MCN}{\datanumber}\sum_{i=1}^{\datanumber}\bigl(\estimatedfunctionnew(\inputvector_i)-\functiontruei\bigr)^2+1.01\E_{(\inputvector,\outputy)}[\noise^2]\\
&~~~~~~~+\tempConst\MCN\constA(\kapastarone)^2\Bigl(\frac{2\actlipc}{\nhiddenlayer}\Bigr)^{2\nhiddenlayer}  \sqrt{\frac{\nhiddenlayer^2\log(2\allparameters)\sum_{i=1}^{\datanumber}\norm{\inputvector_i}^4}{\datanumber^2}}\log(2\datanumber)
\end{align*}
\endgroup
with probability at least $1-1/2\datanumber$.

We finally 1.~invoke the inequality of Corollary~\ref{Paramset} to bound the in-sample-prediction error in the above display with probability at least $1-1/2\datanumber$,
 2.~define $\ConstAP:=\tempConst\MCN\constA$ 
and 3.~use the fact that $\operatorname{risk}(\kapastarone\functionnet_{\thetastarone})=\E_{(\inputvector,\outputy)}[\noise^2] $
to get the desired bound with probability at least $1-1/\datanumber$.

\end{proof}
\subsubsection{Proof of Proposition~\ref{Lipt2}}
\begin{proof}
The proof peels the networks into inner and outer subnetworks.
The inner subnetworks of a network~$\functionclass\in\networkset:=\{\functionclass:\parameterv\in\paraspace\}$ are vector-valued functions defined by
\begin{align*}
   \Aplus_{0}\functionclass\ :\ \R^{\inputdimention}&\to\R^{\inputdimention}\\
   \inputvector&\mapsto\Aplus_{0}\functionclass(\inputvector):=\inputvector
\end{align*}
and
\begin{align*}
   \Aplus_{l}\functionclass\ :\ \R^{\inputdimention}&\to\R^{\neuron_{l}}\\
   \inputvector&\mapsto\Aplus_{l}\functionclass(\inputvector):=\activefunctionb^{l}\rbrtbb{\weightmatrix^{l-1}\activefunctionb^{{l-1}}\rbrtb{\dots\weightmatrix^{1}\activefunctionb^{{1}}\rbrt{\weightmatrix^{0}\inputvector}}}
\end{align*}
for~$l\in\{1,\dots,\nhiddenlayer\}$.
Similarly,
the outer subnetworks of~$\functionclass$ are real-valued functions defined by
\begin{align*}
   \Aminus^{l}\functionclass\ :\ \R^{\neuron_{l-1}}&\to\R\\
   \boldsymbol{z}&\mapsto\Aminus^{l}\functionclass(\boldsymbol{z}):=\weightmatrix^\nhiddenlayer\activefunctionb^{\nhiddenlayer}\rbrtb{\dots\weightmatrix^{l}\activefunctionb^{{l}}\rbrt{\weightmatrix^{l-1}\boldsymbol{z}}}
\end{align*}
for~$l\in\{1,\dots,\nhiddenlayer\}$ and
\begin{align*}
   \Aminus^{\nhiddenlayer+1}\functionclass\ :~\R^{\nhiddenlayer}&\to\R\\
   \boldsymbol{z}&\mapsto\Aminus^{\nhiddenlayer+1}\functionclass(\boldsymbol{z}):=\weightmatrix^\nhiddenlayer\boldsymbol{z}.
\end{align*}

The initial network can be split into an inner and an outer network along every layer~$l\in\cbrt{1,\ldots,\nhiddenlayer+1}$:
 \begin{equation*}\label{eq:dnn}
    \functionclass(\inputvector)=\Aminus^{l}\functionclass\rbrtb{\Aplus_{l-1}\functionclass(\inputvector)}\,.
\end{equation*}
This observation is the basis for the following derivations. 

We now show a contraction property for the inner subnetworks and a Lipschitz property for  the outer subnetworks.
Using the assumption that~$\boldsymbol{z}\mapsto\activefunctionb^{l-1}(\boldsymbol{z})$ is~$\actlipc$-Lipschitz,
we get for every~$\Theta=(\weightmatrix^{\nhiddenlayer},\dots,\weightmatrix^0)$ and~$\inputvector\in \R^{\inputdimention}$ that 
\begingroup
\allowdisplaybreaks
\begin{align*}
 \norm{\Aplus_{l-1}\functionclass(\inputvector)} 
 &=\norm{\activefunctionb^{l-1}\rbrtb{\weightmatrix^{l-2}\Aplus_{l-2}\functionclass(\inputvector)}}\\
 &\leq \actlipc
 \norm{\weightmatrix^{l-2}\Aplus_{l-2}\functionclass(\inputvector)}\\
 &\leq \actlipc \norm{\weightmatrix^{l-2}}\norm{\Aplus_{l-2}\functionclass(\inputvector)} \\
 &\leq \dots\\
 &\leq (\actlipc)^{l-1} \norm{\inputvector}\prod_{j=0}^{l-2}\norm{\weightmatrix^j}
\end{align*}
\endgroup
for all~$l\in\cbrt{2,\ldots,\nhiddenlayer+1}$; 
and one can verify readily that~$\norm{\Aplus_{0}\functionclass(\inputvector)}=\norm{\inputvector}$.
In other words,
$\inputvector\mapsto\Aplus_{l-1}\functionclass(\inputvector)$ and~$\inputvector\mapsto\Aplus_{0}\functionclass(\inputvector)$ are ``contractions'' with constants~$(\actlipc)^{l-1}\prod_{j=0}^{l-2}\norm{W^j}$ and~$1$, respectively,
with respect to the Euclidean norms on the input space~$\R^{\inputdimention}$ and output spaces~$\R^{p_{l-1}}$
 and~$\R^{\inputdimention}$, respectively.

By similar arguments,
we get for every~$\boldsymbol{z}_1,\boldsymbol{z}_2 \in 
\R^{\neuron_{l}}$ that
\begingroup
\allowdisplaybreaks
\begin{align*}
 &|\Aminus^{l+1}\functionclass(\boldsymbol{z}_1)-\Aminus^{l+1}\functionclass(\boldsymbol{z}_2)|\\
 &=\bigl|\weightmatrix^\nhiddenlayer\activefunctionb^\nhiddenlayer\rbrtb{\dots\weightmatrix^{l+1}\activefunctionb^{l+1}(\weightmatrix^l \boldsymbol{z}_1)}-\weightmatrix^\nhiddenlayer\activefunctionb^\nhiddenlayer\rbrtb{\dots\weightmatrix^{l+1}\activefunctionb^{l+1}(\weightmatrix^l \boldsymbol{z}_2)}\bigr|
 \\&\leq \norm{\weightmatrix^\nhiddenlayer}\norm{\activefunctionb^\nhiddenlayer\rbrtb{\dots\weightmatrix^{l+1}\activefunctionb^{l+1}(\weightmatrix^l \boldsymbol{z}_1)}-\activefunctionb^\nhiddenlayer\rbrtb{\dots\weightmatrix^{l+1}\activefunctionb^{l+1}(\weightmatrix^l \boldsymbol{z}_2)}}\\
 &\leq \actlipc\norm{\weightmatrix^\nhiddenlayer}\norm{\weightmatrix^{\nhiddenlayer-1}\activefunctionb^{\nhiddenlayer-1}\rbrtb{\dots\weightmatrix^{l+1}\activefunctionb^{l+1}(\weightmatrix^l \boldsymbol{z}_1)}-\weightmatrix^{\nhiddenlayer-1}\activefunctionb^{\nhiddenlayer-1}\rbrtb{\dots\weightmatrix^{l+1}\activefunctionb^{l+1}(\weightmatrix^l \boldsymbol{z}_2)}}\\
 &\leq \dots \\
 &\leq  (\actlipc)^{L-l}
 \norm{\boldsymbol{z}_1-\boldsymbol{z}_2}\prod_{j=l}^{\nhiddenlayer}\norm{\weightmatrix^j}
\end{align*}
\endgroup
for~$l\in\cbrt{0,\ldots,\nhiddenlayer}$.
In other words,
$\boldsymbol{z}\mapsto\Aminus^{l+1}\functionclass(\mathbf{\boldsymbol{z}})$ is Lipschitz with constant $(\actlipc)^{L-l}\prod_{j=l}^{\nhiddenlayer}\norm{\weightmatrix^j}$ with respect to the Euclidean norms on the input space~$\R^{\neuron_{l}}$ and output space~$\R$.

We now use these contraction and Lipschitz properties for the subnetworks to derive a Lipschitz property for the entire network.
We consider two networks~$\functionclass$ and~$\functionnet_{\Gamma}$ with parameters~$\parameterv=(\wtheta^\nhiddenlayer,\dots,\wtheta^0)\in\paraspace$ and~$\Gamma=(\wgamma^{L},\dots,\wgamma^{0}) \in\paraspace$, respectively. 
    Our above splitting of the  networks 
    applied to~$l=1$ and~$l=\nhiddenlayer+1$
    and the fact that~$\Aplus_0\functionclass(\inputvector)=\Aplus_0\functionnet_{\Gamma}(\inputvector)=\inputvector$
    yield
   \begin{align*}
     |{\functionnet_{\parameterv}(\inputvector)-\functionnet_{{\Gamma}}(\inputvector)}| 
    &=\bigl|\Aminus^1\functionclass\rbrtb{\Aplus_0\functionclass(\inputvector)}-\Aminus^{\nhiddenlayer+1}\functionnet_{\Gamma}\rbrtb{\Aplus_{\nhiddenlayer}\functionnet_{\Gamma}(\inputvector)}\bigr|\\
    &=\bigl|\Aminus^1\functionclass\rbrtb{\Aplus_0\functionnet_{\Gamma}(\inputvector)}-\Aminus^{\nhiddenlayer+1}\functionnet_{\Gamma}\rbrtb{\Aplus_{\nhiddenlayer}\functionnet_{\Gamma}(\inputvector)}\bigr|.  
   \end{align*}
   Elementary algebra and 
   the fact that~$\Aminus^{l+1}\functionclass\rbrt{\Aplus_{l}\functionnet_{\Gamma}(\inputvector)}=\Aminus^{l+1}\functionclass\rbrt{\activefunctionb^{l}\rbrt{\wgamma^{l-1}\Aplus_{l-1}\functionnet_{\Gamma}(\inputvector)}}=\Aminus^{l+2}\functionclass\rbrt{\activefunctionb^{l+1}\rbrt{\wtheta^{l}\Aplus_{l}\functionnet_{\Gamma}(\inputvector)}}$ then allow us to derive
   \begingroup
   \allowdisplaybreaks
   \begin{align*}
    &|{\functionnet_{\parameterv}(\inputvector)-\functionnet_{{\Gamma}}(\inputvector)}| \\
    &=\biggl|\Aminus^1\functionclass\rbrtb{\Aplus_0\functionnet_{\Gamma}(\inputvector)}-\sum_{l=1}^{\nhiddenlayer-1}\rbrtbb{\Aminus^{l+1}\functionclass\rbrtb{\Aplus_{l}\functionnet_{\Gamma}(\inputvector)}-\Aminus^{l+1}\functionclass\rbrtb{\Aplus_{l}\functionnet_{\Gamma}(\inputvector)}}\\
    &~~~~-\rbrtbb{\Aminus^{\nhiddenlayer+1}\functionclass\rbrtb{\Aplus_{\nhiddenlayer}\functionnet_{\Gamma}(\inputvector)}-\Aminus^{\nhiddenlayer+1}\functionclass\rbrtb{\Aplus_{\nhiddenlayer}\functionnet_{\Gamma}(\inputvector)}}-\Aminus^{\nhiddenlayer+1}\functionnet_{\Gamma}\rbrtb{\Aplus_{\nhiddenlayer}\functionnet_{\Gamma}(\inputvector)}\biggr|\\
    &=\biggl|\Aminus^2\functionclass\rbrtbb{\activefunctionb^1\rbrtb{\wtheta^{0}\Aplus_0\functionnet_{\Gamma}(\inputvector)}}\\
    &~~~~-\sum_{l=1}^{\nhiddenlayer-1}\rbrtbbb{\Aminus^{l+1}\functionclass\rbrtbb{\activefunctionb^{l}\rbrtb{\wgamma^{l-1}\Aplus_{l-1}\functionnet_{\Gamma}(\inputvector)}}-\Aminus^{l+2}\functionclass\rbrtbb{\activefunctionb^{l+1}\rbrtb{\wtheta^{l}\Aplus_{l}\functionnet_{\Gamma}(\inputvector)}}}\\
    &~~~~-\Aminus^{\nhiddenlayer+1}\functionclass\rbrtbb{\activefunctionb^{\nhiddenlayer}\rbrtb{\wgamma^{\nhiddenlayer-1}\Aplus_{\nhiddenlayer-1}\functionnet_{\Gamma}(\inputvector)}}+\Aminus^{\nhiddenlayer+1}\functionclass\rbrtb{\Aplus_{\nhiddenlayer}\functionnet_{\Gamma}(\inputvector)}-\Aminus^{\nhiddenlayer+1}\functionnet_{\Gamma}\rbrtb{\Aplus_{\nhiddenlayer}\functionnet_{\Gamma}(\inputvector)}\biggr|\\
    &=\biggl|\sum_{l=1}^{\nhiddenlayer}\rbrtbbb{\Aminus^{l+1}\functionclass\rbrtbb{\activefunctionb^{l}\rbrtb{\wtheta^{l-1}\Aplus_{l-1}\functionnet_{\Gamma}(\inputvector)}}-\Aminus^{l+1}\functionclass\rbrtbb{\activefunctionb^{l}\rbrtb{\wgamma^{l-1}\Aplus_{l-1}\functionnet_{\Gamma}(\inputvector)}}}\\
    &~~~~+\Aminus^{\nhiddenlayer+1}\functionclass\rbrtb{\Aplus_{\nhiddenlayer}\functionnet_{\Gamma}(\inputvector)}-\Aminus^{\nhiddenlayer+1}\functionnet_{\Gamma}\rbrtb{\Aplus_{\nhiddenlayer}\functionnet_{\Gamma}(\inputvector)}\biggr|\\
    &=\biggl|\sum_{l=1}^{\nhiddenlayer}\rbrtbbb{\Aminus^{l+1}\functionclass\rbrtbb{\activefunctionb^{l}\rbrtb{\wtheta^{l-1}\Aplus_{l-1}\functionnet_{\Gamma}(\inputvector)}}-\Aminus^{l+1}\functionclass\rbrtbb{\activefunctionb^{l}\rbrtb{\wgamma^{l-1}\Aplus_{l-1}\functionnet_{\Gamma}(\inputvector)}}}\\
    &~~~~+\wtheta^{\nhiddenlayer}{\Aplus_{\nhiddenlayer}\functionnet_{\Gamma}(\inputvector)}-\wgamma^{\nhiddenlayer}{\Aplus_{\nhiddenlayer}\functionnet_{\Gamma}(\inputvector)}\biggr|\\
    &\le\sum_{l=1}^{\nhiddenlayer}\Bigl|{\Aminus^{l+1}\functionclass\rbrtbb{\activefunctionb^{l}\rbrtb{\wtheta^{l-1}\Aplus_{l-1}\functionnet_{\Gamma}(\inputvector)}}-\Aminus^{l+1}\functionclass\rbrtbb{\activefunctionb^{l}\rbrtb{\wgamma^{l-1}\Aplus_{l-1}\functionnet_{\Gamma}(\inputvector)}}}\Bigr|\\
    &~~~~+\bigl|(\wtheta^{\nhiddenlayer}-\wgamma^{\nhiddenlayer}){\Aplus_{\nhiddenlayer}\functionnet_{\Gamma}(\inputvector)}\bigr|.
\end{align*}
\endgroup
We bound this further by using 1.~the above-derived Lipschitz property of~$\Aminus^{l+1}\functionclass$, 2.~the assumption that the~$\activefunctionb^l$ are~$\actlipc$-Lipschitz,
 3.~the properties of the $\ell_2$-norm,
 and 4.~the above-derived contraction property of~$\Aplus_{l-1}\functionnet_\Gamma$:
\begingroup
\allowdisplaybreaks
\begin{align*}
    &|{\functionnet_{\parameterv}(\inputvector)-\functionnet_{{\Gamma}}(\inputvector)}|\\
    &\leq\sum_{l=1}^{\nhiddenlayer}\biggl[(\actlipc)^{L-l}\prod_{j=l}^{\nhiddenlayer}{\norm{\weightmatrix^j}\biggr]}\norm{\activefunctionb^{l}\rbrtb{\wtheta^{l-1}\Aplus_{l-1}\functionnet_\Gamma(\inputvector)}-\activefunctionb^{l}\rbrtb{\wgamma^{l-1}\Aplus_{l-1}\functionnet_\Gamma(\inputvector)}}\\
    &~~~~+\bigl|(\wtheta^{\nhiddenlayer}-\wgamma^{\nhiddenlayer}){\Aplus_{\nhiddenlayer}\functionnet_{\Gamma}(\inputvector)}\bigr|\\
     &\leq\sum_{l=1}^{\nhiddenlayer}\biggl[(\actlipc)^{L-l+1}\prod_{j=l}^{\nhiddenlayer}{\norm{\weightmatrix^j}\biggr]}\norm{\wtheta^{l-1}\Aplus_{l-1}\functionnet_\Gamma(\inputvector)-\wgamma^{l-1}\Aplus_{l-1}\functionnet_\Gamma(\inputvector)}+\bigr|(\wtheta^{\nhiddenlayer}-\wgamma^{\nhiddenlayer}){\Aplus_{\nhiddenlayer}\functionnet_{\Gamma}(\inputvector)}\bigl|
     \\
    &\leq\sum_{l=1}^{\nhiddenlayer}\biggl[(\actlipc)^{L-l+1}\prod_{j=l}^{\nhiddenlayer}{\norm{\weightmatrix^j}\biggr]}\norm{\wtheta^{l-1}-\wgamma^{l-1}}\norm{\Aplus_{l-1}\functionnet_\Gamma(\inputvector)}+\norm{\wtheta^{\nhiddenlayer}-\wgamma^{\nhiddenlayer}}\norm{\Aplus_{\nhiddenlayer}\functionnet_{\Gamma}(\inputvector)}
    \\
    &\leq\sum_{l=1}^{\nhiddenlayer}\biggl[(\actlipc)^{L-l+1}\prod_{j=l}^{\nhiddenlayer}{\norm{\weightmatrix^j}\biggr]}\norm{\wtheta^{l-1}-\wgamma^{l-1}}\sbrtbbb{(\actlipc)^{l-1}\prod_{j=0}^{l-2}\norm{\wgamma^j}}\norm{\inputvector}\\
    &~~~~+\norm{\wtheta^{\nhiddenlayer}-\wgamma^{\nhiddenlayer}}\sbrtbbb{(\actlipc)^{L}\prod_{j=0}^{\nhiddenlayer-1}\norm{\wgamma^j}}\norm{\inputvector},
\end{align*}
\endgroup
where we set~$\prod_{j=0}^{-1}\norm{\wgamma^j}:=1$.
Consolidating and rearranging then yields
\begingroup
\allowdisplaybreaks
\begin{align*}
    &|{\functionnet_{\parameterv}(\inputvector)-\functionnet_{{\Gamma}}(\inputvector)}|\\
    &\leq(\actlipc)^{L}\Biggl(\sum_{l=1}^{\nhiddenlayer}\biggl[
    {\prod_{\substack{j\in\{0,\dots,\nhiddenlayer\}\\j\neq l-1}}}{\rbrtb{\norm{\weightmatrix^j}\vee\norm{\wgamma^j}}\biggr]}\norm{\wtheta^{l-1}-\wgamma^{l-1}}\norm{\inputvector}\\
    &~~~~~~~~~~~~~~~+\sbrtbbb{\prod_{j=0}^{\nhiddenlayer-1}\rbrtb{\norm{\weightmatrix^j}\vee\norm{\wgamma^j}}}\norm{\wtheta^{\nhiddenlayer}-\wgamma^{\nhiddenlayer}}\norm{\inputvector}\Biggr)\\
    &=(\actlipc)^{L}\sum_{l=1}^{\nhiddenlayer+1}\biggl[{\prod_{\substack{j\in\{0,\dots,\nhiddenlayer\}\\j\neq l-1}}}{\rbrtb{\norm{\weightmatrix^j}\vee\norm{\wgamma^j}}\biggr]}\norm{\wtheta^{l-1}-\wgamma^{l-1}}\norm{\inputvector}
    \\
    &\le(\actlipc)^{L}\norm{\inputvector}\max_{l\in\cbrt{1,\dots,\nhiddenlayer+1}}\cbrtbbb{{\prod_{\substack{j\in\{0,\dots,\nhiddenlayer\}\\j\neq l-1}}}\rbrtb{\norm{\weightmatrix^j}\vee\norm{\wgamma^j}}}\sum_{m=1}^{\nhiddenlayer+1}\norm{\wtheta^{m-1}-\wgamma^{m-1}}
    \\
    &=(\actlipc)^{L}\norm{\inputvector}\max_{l\in\cbrt{0,\dots,\nhiddenlayer}}\cbrtbbb{{\prod_{\substack{j\in\{0,\dots,\nhiddenlayer\}\\j\neq l}}}\rbrtb{\norm{\weightmatrix^j}\vee\norm{\wgamma^j}}}\sum_{m=0}^{\nhiddenlayer}\norm{\wtheta^{m}-\wgamma^{m}}.
\end{align*}
\endgroup

We now study the last sum in that bound:
First, we observe that
\begin{align*}
  \sum_{m=0}^{L}\norm{\wtheta^{m}-\wgamma^{m}}&=\sqrt{\rbrtbbb{\sum_{m=0}^{L}\norm{\wtheta^{m}-\wgamma^{m}}}^2}\\
  &\le\sqrt{(L+1)\sum_{m=0}^{L}\norm{\wtheta^{m}-\wgamma^{m}}^2}\\
  &=\sqrt{L+1}\sqrt{\sum_{m=0}^{L}\norm{\wtheta^{m}-\wgamma^{m}}^2},
\end{align*}
where we use~$\rbrtb{\sum_{m=0}^{\nhiddenlayer}a_{m}}^2\le (\nhiddenlayer+1) \sum_{m=0}^{\nhiddenlayer}(a_{m})^2$ with~$a_{m}:=\norm{\wtheta^{m}-\wgamma^{m}}$. 
We then bound the last line further to obtain
\begin{align*}
   \sum_{m=0}^{L}\norm{\wtheta^{m}-\wgamma^{m}}
  &\le\sqrt{L+1}\normP{\parameterv-\Gamma}\\
   &\leq2\sqrt{L}\normP{\parameterv-\Gamma}
   \\
   &\leq2\sqrt{L}\normFP{\parameterv-\Gamma},
\end{align*}
where we use 1.~the definition of the operator norm on  Page~\pageref{Fnormmatrix}, 2.~$\sqrt{1+\nhiddenlayer}\leq  2\sqrt{\nhiddenlayer}$, and 3.~$\normP{\parameterv-\Gamma}\leq \normFP{\parameterv-\Gamma}$.
Combining this result with the previous display yields
\begin{align*}    |{\functionnet_{\parameterv}(\inputvector)-\functionnet_{{\Gamma}}(\inputvector)}|
    &\leq2(\actlipc)^{L}\sqrt{\nhiddenlayer}\norm{\inputvector}\max_{l\in\cbrt{0,\dots,\nhiddenlayer}}\cbrtbbb{{\prod_{\substack{j\in\{0,\dots,\nhiddenlayer\}\\j\neq l}}}\rbrtb{\norm{\weightmatrix^j}\vee\norm{\wgamma^j}}}\normFP{\parameterv-\Gamma}\\
    &=\Lipfunc\normFP{\parameterv-\Gamma},
\end{align*}
as desired.

The second claim then follows readily:
\begin{align*}
    \normfcn{\functionnet_{\parameterv}-\functionnet_{{\Gamma}}}_\datanumber&=\sqrt{\frac{1}{\datanumber}\sum_{i=1}^{\datanumber}\rbrtb{\functionnet_{\parameterv}(\inputvector_i)-\functionnet_{{\Gamma}}(\inputvector_i)}^2}\\
    &\leq\sqrt{\frac{1}{\datanumber}\sum_{i=1}^{\datanumber}\rbrtb{\LipfuncN(\inputvector_i)\normFP{\parameterv-\Gamma}}^2}\\
    &=\sqrt{\frac{1}{\datanumber}\sum_{i=0}^{\datanumber}\rbrtbb{2(\actlipc)^{L}\sqrt{\nhiddenlayer}\norm{\inputvector_i}\max_{l\in\cbrt{0,\dots,\nhiddenlayer}}\cbrtbb{\prod_{\substack{j\in\cbrt{0,\dots,\nhiddenlayer}\\j\neq l}}\rbrtb{\norm{\weightmatrix^j}\vee\norm{\wgamma^j}}}\normFP{\parameterv-\Gamma}}^2}\\
       &=2(\actlipc)^{L}\sqrt{\nhiddenlayer}\sqrt{\frac{1}{\datanumber}\sum_{i=0}^{\datanumber}\norm{\inputvector_i}^2}\max_{l\in\cbrt{0,\dots,\nhiddenlayer}}\cbrtbb{\prod_{\substack{j\in\cbrt{0,\dots,\nhiddenlayer}\\j\neq l}}\rbrtb{\norm{\weightmatrix^j}\vee\norm{\wgamma^j}}}\normFP{\parameterv-\Gamma}\\
    &=\LipConst\normFP{\parameterv-\Gamma},
\end{align*}
as desired.
\end{proof}
\subsubsection{Proof of Lemma~\ref{UPfunA1}}
\begin{proof}
 The proof follows from Proposition~\ref{Lipt2} and restricting the parameter space to $\paraspaceN$. 
Since $\thetasym,\Gamma\in\paraspaceN$, we can get
  \begin{align*}
  \sum_{j=0}^{\nhiddenlayer}\rbrtb{\norm{\weightmatrix^j}\vee\norm{\wgamma^j}}&\le  \sum_{j=0}^{\nhiddenlayer}\rbrtb{\norm{\weightmatrix^j}+\norm{\wgamma^j}}\\
  &\le \sum_{j=0}^{\nhiddenlayer}\rbrtb{\normone{\weightmatrix^j}+\normone{\wgamma^j}}\\
  &=\normoneP{\thetasym}+\normoneP{\Gamma}\\
  &\le 2.
   \end{align*}
   Using 1.~the inequality of arithmetic and geometric means, 
2.~the nonnegativity of norms ($\norm{\cdot}\ge 0$),  
and 3.~the above display,
 we obtain
   \begin{align*}
   \max_{l\in\cbrt{0,\dots,\nhiddenlayer}} \prod_{\substack{j\in\cbrt{0,\dots,\nhiddenlayer}\\j\neq l}}\rbrtb{\norm{\weightmatrix^j}\vee\norm{\wgamma^j}}&\le  \max_{l\in\cbrt{0,\dots,\nhiddenlayer}}  \Biggl(\frac{1}{\nhiddenlayer} \sum_{j=0,j\neq l}^{\nhiddenlayer}\rbrtb{\norm{\weightmatrix^j}\vee\norm{\wgamma^j}}\Biggr)^\nhiddenlayer\\
   &\le \Biggl(\frac{1}{\nhiddenlayer} \sum_{j=0}^{\nhiddenlayer}\rbrtb{\norm{\weightmatrix^j}\vee\norm{\wgamma^j}}\Biggr)^\nhiddenlayer\\
   &\le \Bigl(\frac{2}{\nhiddenlayer}\Bigr)^\nhiddenlayer.
   \end{align*}
We can plug this inequality into the definition of~$\LipConst$  in Proposition~\ref{Lipt2} to get
   \begin{equation*}
  \LipConstone=2(\actlipc)^\nhiddenlayer\sqrt{\nhiddenlayer}\normfcn{\inputvector}_{\datanumber} \Bigl(\frac{2}{\nhiddenlayer}\Bigr)^\nhiddenlayer,
   \end{equation*}
   as desired in the first claim.
   The second claim then follows by setting~$\Gamma$ equal to the all-zeros parameter in the first claim.
\end{proof}

\subsubsection{Proof of Proposition~\ref{EN}}
\begin{proof}
We prove  the  two  claims  in  order.

\emph{Claim 1:} entropy bound

Our strategy is to move from~$\entropy({\radiusu,\networksetN,\normfcn{\cdot}_{\datanumber})}$ 
to~$\entropy(\radiusu/\LipConstone,{\paraspaceN},\normFP{\cdot})$ via Proposition~\ref{Lipt2} and Lemma~\ref{generalLP} and then bound the latter covering number  using a bound on the entropy of $\ell_1$-balls.

Lemma~\ref{UPfunA1} ensures that the function~$\thetasym\mapsto\function_{\thetasym}$ restricted to the parameter space~$\paraspaceN$ is~$\LipConstone$-Lipschitz with respect to the prediction distance~$\normfcn{\cdot}_{\datanumber}$ on the network space and Frobenius norm~$\normFP{\cdot}$ on the parameter space with~$\LipConstone=2(2\actlipc/L)^\nhiddenlayer\sqrt{\nhiddenlayer}\normfcn{\inputvector}_{\datanumber}$. 
If~$\LipConstone=0$, then~$\covernpung(\radiusu,\networksetN,\normfcn{\cdot}_\datanumber)=1$ for all $r\in(0,\infty)$  and, therefore,  $\entropy\rbrt{\radiusu,\networksetN,\normfcn{\cdot}_{\datanumber}}=0$ for all $r\in(0,\infty)$, which is commensurate with the alleged bound.
We can thus assume $\LipConstone>0$ in the following.

Since $\normfcn{\functionnet_\Gamma-\functionnet_{\boldsymbol{0}_\paraspace}}_{\datanumber}=\normfcn{\functionnet_\Gamma}_\datanumber\leq  \sup_{\thetasym\in{\paraspaceN} }\normfcn{\functionnet_\thetasym}_{\datanumber}=:\uppboundfunc$ for all~$\Gamma\in\paraspaceN$ and~$\boldsymbol{0}_{\paraspace}:=(\boldsymbol{0}_{p_{\nhiddenlayer+1}\times p_\nhiddenlayer},\dots,\boldsymbol{0}_{p_{1}\times p_0})$,
it holds that~$\covernpung(\radiusu,\networksetN,\normfcn{\cdot}_\datanumber)=1$ for all~$\radiusu>\uppboundfunc$ and, consequently, $\entropy(\radiusu,\networksetN,\normfcn{\cdot}_\datanumber)=0$ for all~$\radiusu>\uppboundfunc$, which is commensurate with the alleged bound. We can thus assume $\radiusu\le\uppboundfunc$ in the following. 

We then apply Lemma~\ref{generalLP}
with~$\paraspaceprime:=\paraspaceN$,~$\networksetprime:=\networksetN$,~$\normfcn{\Lipfcng}_{\datanumber}:=\LipConstone$, and~$\parammetric:=\normFP{\cdot}$ to obtain 
\begin{equation*}
  \entropy\rbrtb{\radiusu,\networksetN,\normfcn{\cdot}_{\datanumber}}\leq  \entropy\rbrtbb{\frac{\radiusu}{\LipConstone},{\paraspaceN},\normFP{\cdot}}.
\end{equation*}

\newcommand{\newomegav}{\boldsymbol{\omega}}
\newcommand{\newomegas}{\omega}
We now think of~$\paraspaceN$
as a set in~$\R^{\allparameters}$.
Defining~$\paraspaceprimetwo:=\cbrt{\newomegav=(\newomegas_1,\dots,\newomegas_{\allparameters})\tp\in\R^{\allparameters}\,:\, \sum_{j=1}^{\allparameters}|\newomegas_j|\leq 1}$,
we find for every~$\radiusu\le\uppboundfunc$  and~$\radiuseta:= \radiusu/(\sqrt{2}\LipConstone)\in(0,1)$, where $\radiuseta\in (0,1)$ comes by the definition of $\radiuseta$ together with $\radiusu\le\uppboundfunc$ and $\uppboundfunc\leq\LipConstone$ (by Lemma~\ref{UPfunA1}), 
\begin{equation*}
    \entropy\rbrtbb{\frac{\radiusu}{\LipConstone},\paraspaceN,\normFP{\cdot}}=\entropy\rbrtb{\sqrt{2}\radiuseta,\paraspaceprimetwo,\norm{\cdot}}.
\end{equation*}

We then bound the right-hand side of this equality using 1.~the definition of the entropy, 2.~the entropy bound of~\citet[Page~9]{lederer2010bounds} (with $k=\lceil2nA^2M^2/\radiuseta^2\rceil$, $M=1$, and $A=1/\sqrt{n}$), and 3.~a simplification (by $\radiuseta\in (0,1)$)  to get
\begin{align*}
\entropy\rbrtb{\sqrt{2}\radiuseta,\paraspaceprimetwo,\norm{\cdot}}&=\log \covernpung\rbrtb{\sqrt{2}\radiuseta,\paraspaceprimetwo,\norm{\cdot}} \\
&\le \rbrtbb{\frac{2}{\radiuseta^2}+1}\log\bigl(e(1+\allparameters\radiuseta^2)\bigr)\\
&\le
\frac{3}{\radiuseta^2}\log(2e\allparameters\radiuseta^2\vee 2e).
\end{align*}
Collecting the pieces and recalling that~$\radiuseta= \radiusu/(\sqrt{2}\LipConstone)$ then yields
\begin{align*}\label{ENT0}
  \entropy\rbrtb{\radiusu,\networksetN,\normfcn{\cdot}_{\datanumber}}
  &\leq\entropy\rbrtb{\sqrt{2}\radiuseta,\paraspaceprimetwo,\norm{\cdot}}\\
  &\leq \frac{6(\LipConstone)^2 }{\radiusu^2}{\log\rbrtbbb{\frac{e\allparameters\radiusu^2}{(\LipConstone)^2}\vee 2e }},
\end{align*}
as desired.

\emph{Claim 2:} Dudley bound

 Our strategy is to use  \emph{Claim~1} to prove that
 \begin{equation*}
  \entropyboundN\leq \frac{5\LipConstone}{2} \sqrt{\log{(e\allparameters\vee 2e })}\log\rbrtbb{\frac{8\sigma\uppboundfunc}{\radiusdelta}}
 \end{equation*}
 and then to use  
  Lemma~\ref{UPfunA1} to formulate the bound in the desired way.

We first split the Dudley integral into two parts according to
 \begin{equation*}
   \entropyboundN
 = \int_{\radiusdelta/(8\sigma)}^{\uppboundfunc}{\entropy^{1/2}\bigl(\radiusu,\networksetN,\normfcn{\cdot}_\datanumber\bigr)d\radiusu}+ \int_{\radiusu>\uppboundfunc}{\entropy^{1/2}\bigl(\radiusu,\networksetN,\normfcn{\cdot}_\datanumber\bigr)d\radiusu}.
 \end{equation*}

Recalling $\entropy\bigl(\radiusu,\networksetN,\normfcn{\cdot}_\datanumber\bigr)=0$ for all~$\radiusu>\uppboundfunc$, 
the Dudley integral simplifies to
\begin{equation*}
   \entropyboundN
 = \int_{\radiusdelta/(8\sigma)}^{\uppboundfunc}{\entropy^{1/2}\bigl(\radiusu,\networksetN,\normfcn{\cdot}_\datanumber\bigr)d\radiusu}. 
\end{equation*}

 Using this equality together with the bound from \emph{Claim~1}, we obtain that
\begin{align*}
 \entropyboundN
 &= \int_{\radiusdelta/(8\sigma)}^{\uppboundfunc}{\entropy^{1/2}\bigl(\radiusu,\networksetN,\normfcn{\cdot}_\datanumber\bigr)d\radiusu}  \\
&\le\int_{\radiusdelta/(8\sigma)}^{\uppboundfunc}\Biggl(\frac{6(\LipConstone)^2 }{\radiusu^2}{\log\rbrtbbb{\frac{e\allparameters\radiusu^2}{(\LipConstone)^2} \vee 2e}}\Biggr)^{1/2} d\radiusu \\
&\leq  \frac{5\LipConstone}{2} \sqrt{\log{\rbrtbbb{\frac{e\allparameters\uppboundfunc^2}{(\LipConstone)^2}\vee 2e }}}\int_{\radiusdelta/(8\sigma)}^{\uppboundfunc}\frac{1}{\radiusu}d\radiusu\\
&=  \frac{5\LipConstone}{2} \sqrt{\log{\rbrtbbb{\frac{e\allparameters\uppboundfunc^2}{(\LipConstone)^2}\vee 2e }}}\log\rbrtbb{\frac{8\sigma\uppboundfunc}{\radiusdelta}}.
\end{align*}
Since~$\uppboundfunc\leq\LipConstone$ by  Lemma~\ref{UPfunA1}, we can  get
\begin{equation*}
    \entropyboundN
 \leq  \frac{5\LipConstone}{2} \sqrt{\log{(e\allparameters\vee 2e })}\log\rbrtbb{\frac{8\sigma\LipConstone}{\radiusdelta}},
\end{equation*}
as desired.
\end{proof}
\subsubsection{Proof of Lemma~\ref{generalLP}}
\begin{proof}
The case~$\normfcn{\Lipfcng}_{\datanumber}=0$ follows directly from our convention~$a/0=\infty$ for~$a\in(0,\infty)$ on Page~\pageref{generalLP} and the definition of the entropy on Page~\pageref{entropyint}.
We thus assume~$\normfcn{\Lipfcng}_{\datanumber}>0$ in the following.

Using the definition of the prediction distance on Page~\pageref{empnorm} and the  Lipschitz property stipulated in the lemma, we find that
\begingroup
\allowdisplaybreaks
      \begin{align*}
        \normfcn{\functionnet_{{\parameterv}}-\functionnet_{\Gamma}}_{\datanumber} &=\sqrt{\frac{1}{\datanumber}\sum_{i=1}^{\datanumber}\bigl(\functionnet_{\parameterv}(\inputvector_{i})-\functionnet_{\Gamma}(\inputvector_{i})\bigr)^2}\\
        &\leq \sqrt{\frac{1}{\datanumber}\sum_{i=1}^{\datanumber}{\big(\Lipfcng(\inputvector_i)\big)}^2{\big(\parammetric({\parameterv,{\Gamma})\bigr)}}^2}\\
         &= \normfcn{\Lipfcng}_{\datanumber}\parammetric(\parameterv,{\Gamma}).
      \end{align*}
\endgroup
      
Now,
  let~$\paramcovering$
  be an~$\radiusu/\normfcn{\Lipfcng}_{\datanumber}$-covering of~$\paraspaceprime$ with respect to the metric~$\parammetric$. 
  This means that  for every~$\parameterv\in \paraspaceprime$, there is a~$\parameterveps\in \paramcovering$  such that~$\parammetric(\parameterv,\parameterveps)\leq\radiusu/\normfcn{\Lipfcng}_{\datanumber}$. 
 This insight together with the first display applied to~$\Gamma=\parameterveps$ yield that for every function~$\functionclass \in \networksetprime$, there is a~$\functionnet_{\parameterveps}\in \cbrt{\functionnet_{\parameterveps}: {\parameterveps}\in \paramcovering}$
  such that  
  \begin{align*}
     \normfcn{\functionclass-\functionnet_{\parameterveps}}_{\datanumber}
     & \leq \normfcn{\Lipfcng}_{\datanumber}\parammetric(\parameterv,\parameterveps)\\
      & \leq \normfcn{\Lipfcng}_{\datanumber}\cdot\frac{\radiusu}{\normfcn{\Lipfcng}_{\datanumber}}\\
      &= \radiusu.
   \end{align*}
   Hence, $\cbrt{\functionnet_{\parameterveps}: {\parameterveps}\in \paramcovering}$ is an~$\radiusu$-covering of~$\networksetprime$ with respect to~$\normfcn{\cdot}_{\datanumber}$. 
   The proof then follows directly from the definition of the entropy on Page~\pageref{entropyint} as the logarithm of the covering number.
\end{proof}

\subsubsection{Proof of Lemma~\ref{subgausslemma}}
\begin{proof}
There are several ways to derive such a deviation inequality.
We choose an approach based on a version of Bernstein's inequality.

A Taylor expansion of the sub-Gaussian assumption on Page~\pageref{subgauss} gives
\begin{equation*}
 \max_{i\in\{1,\dots,\datanumber\}} \noisedis^2\rbrtbb{ \mathbb{E}\bigl[|\noise_i|^2/\noisedis^2+\rbrtb{|\noise_i|^2/\noisedis^2}^2/2!+\rbrtb{|\noise_i|^2/\noisedis^2}^3/3!+\dots\bigr]}\le \subGparaone^2.
\end{equation*}
Hence, the individual terms of the expansion satisfy the moment inequality
\begin{equation*}
  \max_{i\in\{1,\dots,\datanumber\}} \noisedis^2 \mathbb{E}\Bigl[\rbrtb{|\noise_i|^2/\noisedis^2}^m/{m!}\Bigr]\le \subGparaone^2~~~~~~~~~~\text{for~all~}m\in\{1,2,\dots\}.
\end{equation*}
By exchanging the maximum for an average, we then find
\begin{equation*}\frac{1}{\datanumber}\sum_{i=1}^\datanumber \noisedis^2 \mathbb{E}\Bigl[\rbrtb{|\noise_i|^2/\noisedis^2}^m/{m!}\Bigr]\le \subGparaone^2~~~~~~~~~~\text{for~all~} m\in\{1,2,\dots\},
\end{equation*}
which can be reformulated as
\begin{equation*}
  \sum_{i=1}^\datanumber \mathbb{E}\Bigl[\rbrtb{|\noise_i|^2}^m\Bigr]\le \frac{m!}{2} (2\datanumber\subGparaone^2 \noisedis^2) (\noisedis^2)^{m-2} ~~~~~~~~~~\text{for~all~} m\in\{1,2,\dots\}.
\end{equation*}
This means that the squared noise random variables satisfy a ``Bernstein condition''~\citep{Sara2013Orliz}.

We can thus apply a Bernstein-type deviation inequality such as \citet[Corollary~2.11]{Boucheron2013} to derive
\begin{equation*}
  \mathbb{P}\biggl(
  \sum_{i=1}^{\datanumber}\rbrtbb{(\noise_i)^2-\mathbb{E}\bigl[\rbrt{\noise_i}^2\bigr]}\geq\frac{\datanumber\subgausslevel}{2}\biggr)\le e^{-\frac{\datanumber^2 \subgausslevel^2/4}{2\rbrt{2\datanumber\subGparaone^2\noisedis^2+\noisedis^2\datanumber\subgausslevel/2}}},
\end{equation*}
which can be reformulated as
\begin{equation*}
  \mathbb{P}\biggl(
  \frac{1}{\datanumber}\sum_{i=1}^{\datanumber}\rbrtbb{(\noise_i)^2-\mathbb{E}\bigl[\rbrt{\noise_i}^2\bigr]}\geq \frac{\subgausslevel}{2}\biggr)\le e^{-\frac{\datanumber \subgausslevel^2}{16\subGparaone^2\noisedis^2+4\subgausslevel\noisedis^2}}.
\end{equation*}
Using that~$\subgausslevel\geq 2\subGparaone^2$ by assumption,
we then find further
\begin{equation*}
    \mathbb{P}\biggl(
  \frac{1}{\datanumber}\sum_{i=1}^{\datanumber}\rbrtbb{(\noise_i)^2-\mathbb{E}\bigl[\rbrt{\noise_i}^2\bigr]}\geq \frac{\subgausslevel}{2}\biggr)\le e^{-\frac{\datanumber \subgausslevel^2}{8\subgausslevel\noisedis^2+4\subgausslevel\noisedis^2}}=e^{-\frac{\datanumber \subgausslevel}{12\noisedis^2}}.
\end{equation*}

By 1.~adding a zero-valued term, 
2.~invoking the above-derived property on the~$(\noise_i)^2$ (set~$m=1$),
3.~using again that~$\subgausslevel\geq 2\subGparaone^2$,
and 4.~invoking the above display,
we conclude that
\begin{align*}
    \mathbb{P}\biggl(\frac{1}{\datanumber}\sum_{i=1}^{\datanumber}{(\noise_i)^2}\geq\subgausslevel\biggr)&=\mathbb{P}\biggl(\frac{1}{\datanumber}\sum_{i=1}^{\datanumber}{\Bigl((\noise_i)^2}-\mathbb{E}\bigl[(\noise_i)^2\bigr]\Bigr)\geq \subgausslevel-\frac{1}{\datanumber}\sum_{i=1}^{\datanumber}\mathbb{E}\bigl[(\noise_i)^2\bigr]\biggr)\\
    &\leq\mathbb{P}\biggl(\frac{1}{\datanumber}\sum_{i=1}^{\datanumber}{\Bigl((\noise_i)^2}-\mathbb{E}\bigl[(\noise_i)^2\bigr]\Bigr)\geq \subgausslevel-\subGparaone^2\biggr)\\
    &\leq\mathbb{P}\biggl(\frac{1}{\datanumber}\sum_{i=1}^{\datanumber}{\Bigl((\noise_i)^2}-\mathbb{E}\bigl[(\noise_i)^2\bigr]\Bigr)\geq \frac{\subgausslevel}{2}\biggr)\\
    &\leq e^{-\frac{\datanumber \subgausslevel}{12\noisedis^2}},
\end{align*}
as desired.
\end{proof}

\section{Discussion}\label{sec:discussion}
Our theories in Section~\ref{sec:example-l1} show that $\ell_1$-regularization can guarantee accurate prediction even when the neural networks are very wide (see the logarithmic increase of the error in the number of parameters) and deep (see the decrease of the error in the number of layers).
More generally, our theories in Section~\ref{sec:mainresult} facilitate the derivation of concrete guarantees by connecting regularization with the rich literature on suprema of empirical processes.


Another related contribution is \citet{Hieber2017},
which uses ideas from nonparameteric statistics to derive bounds for empirical-risk minimization over classes of sparse networks.
Direct sparsity constraints, in contrast to $\ell_1$-regularization, are not feasible in practice.
But \citet{Hieber2017} provides a number of new insights,
two of which  are also important here:
first, it highlights the statistical benefits of sparsity and, therefore, supports our results in Section~\ref{sec:example-l1};
and second,
it indicates that---arguably under strict assumptions---one can achieve the rate $1/\datanumber$ rather than $1/\sqrt{\datanumber}$.
 However, again, we believe that the $1/\sqrt{n}$-rate cannot be improved in general:
  while a formal proof still needs to be established,
  a corresponding statement has already been proved for $\ell_1$-regularized linear regression~\cite[Proposition~4]{Dalalyan2017}.
 In this sense, we might claim some optimality of our results.

Our theory considers only global minima of the estimators' objective functions,
while the objective functions might also have saddle points or suboptimal local minima.
However, current research suggests that at least for wide networks,
global optimization is feasible---see \citet{Lederer20} and references therein.


In summary,
our paper highlights the effectiveness of regularisation in deep learning, and it furthers the mathematical understanding of neural networks more broadly.
As practical advice, our results suggest the use of wide networks (to minimize approximation errors and to facilitate optimizations) with many layers (to improve statistical accuracy)  together with regularization (to avoid overfitting).

\subsection*{Acknowledgments}
We thank Aleksandr Beknazaryan for his input,
which led to an improvement of our bounds,
and Shih-Ting Huang and Johannes Schmidt-Hieber for their valuable feedback.

 \small
\bibliography{references}

\begin{thebibliography}{45}
\providecommand{\natexlab}[1]{#1}
\providecommand{\url}[1]{\texttt{#1}}
\expandafter\ifx\csname urlstyle\endcsname\relax
  \providecommand{\doi}[1]{doi: #1}\else
  \providecommand{\doi}{doi: \begingroup \urlstyle{rm}\Url}\fi

\bibitem[Anthony and Bartlett(2009)]{Anthony2009}
M.~Anthony and P.~Bartlett.
\newblock \emph{Neural network learning: Theoretical foundations}.
\newblock Cambridge Univ.\@ Press, 2009.

\bibitem[Badrinarayanan et~al.(2017)Badrinarayanan, Kendall, and
  Cipolla]{Badrinarayanan2017}
V.~Badrinarayanan, A.~Kendall, and R.~Cipolla.
\newblock {SegNet}: A deep convolutional encoder-decoder architecture for image
  segmentation.
\newblock \emph{IEEE Trans.\@ Pattern Anal.\@ Mach.\@ Intell.\@}, 39\penalty0
  (12):\penalty0 2481--2495, 2017.

\bibitem[Barron and Klusowski(2018)]{Barron2018}
A.~Barron and J.~Klusowski.
\newblock Approximation and estimation for high-dimensional deep learning
  networks.
\newblock \emph{arXiv:1809.03090}, 2018.

\bibitem[Barron and Klusowski(2019)]{Barron2019}
A.~Barron and J.~Klusowski.
\newblock Complexity, statistical risk, and metric entropy of deep nets using
  total path variation.
\newblock \emph{arXiv:1902.00800}, 2019.

\bibitem[Bartlett(1998)]{Bartlett1998}
P.~Bartlett.
\newblock The sample complexity of pattern classification with neural networks:
  The size of the weights is more important than the size of the network.
\newblock \emph{IEEE Trans.\@ Inform.\@ Theory}, 44\penalty0 (2):\penalty0
  525--536, 1998.

\bibitem[Bartlett and Mendelson(2002)]{Bartlett2002}
P.~Bartlett and S.~Mendelson.
\newblock {Rademacher} and {Gaussian} complexities: risk bounds and structural
  results.
\newblock \emph{J.\@ Mach.\@ Learn.\@ Res.\@}, 3:\penalty0 463--482, 2002.

\bibitem[Boucheron et~al.(2013)Boucheron, Lugosi, and Massart]{Boucheron2013}
S.~Boucheron, G.~Lugosi, and P.~Massart.
\newblock \emph{Concentration inequalities: A nonasymptotic theory of
  independence}.
\newblock Oxford Univ.\@ Press, 2013.

\bibitem[Cand{\`e}s et~al.(2006)Cand{\`e}s, Romberg, and Tao]{Candes2006}
E.~Cand{\`e}s, J.~Romberg, and T.~Tao.
\newblock Stable signal recovery from incomplete and inaccurate measurements.
\newblock \emph{Commun.\@ Pure Appl.\@ Math.\@}, 59\penalty0 (8):\penalty0
  1207--1223, 2006.

\bibitem[Chorowski et~al.(2015)Chorowski, Bahdanau, Serdyuk, Cho, and
  Bengio]{Chorowski2015}
J.~Chorowski, D.~Bahdanau, D.~Serdyuk, K.~Cho, and Y.~Bengio.
\newblock Attention-based models for speech recognition.
\newblock In \emph{{NIPS}}, pages 577--585, 2015.

\bibitem[Clevert et~al.(2015)Clevert, Unterthiner, and
  Hochreiter]{Clevert2015ELU}
D.~Clevert, T.~Unterthiner, and S.~Hochreiter.
\newblock Fast and accurate deep network learning by exponential linear units
  ({ELUs}).
\newblock \emph{arXiv:1511.07289}, 2015.

\bibitem[Dalalyan et~al.(2017)Dalalyan, Hebiri, and Lederer]{Dalalyan2017}
A.~Dalalyan, M.~Hebiri, and J.~Lederer.
\newblock On the prediction performance of the lasso.
\newblock \emph{Bernoulli}, 23\penalty0 (1):\penalty0 552--581, 2017.

\bibitem[Donoho(2006)]{Donoho2006}
D.~Donoho.
\newblock Compressed sensing.
\newblock \emph{IEEE Trans.\@ Inform.\@ Theory}, 52\penalty0 (4):\penalty0
  1289--1306, 2006.

\bibitem[Du et~al.(2018)Du, Hu, and Lee]{Du2018}
S.~Du, W.~Hu, and J.~Lee.
\newblock Algorithmic regularization in learning deep homogeneous models:
  layers are automatically balanced.
\newblock In \emph{NIPS}, pages 384--395, 2018.

\bibitem[Elfwing et~al.(2018)Elfwing, Uchibe, and Doya]{Elfwing2018}
S.~Elfwing, E.~Uchibe, and K.~Doya.
\newblock Sigmoid-weighted linear units for neural network function
  approximation in reinforcement learning.
\newblock \emph{Neural Networks}, 107:\penalty0 3--11, 2018.

\bibitem[Girshick et~al.(2014)Girshick, Donahue, Darrell, and
  Malik]{Girshick2014}
R.~Girshick, J.~Donahue, T.~Darrell, and J.~Malik.
\newblock Rich feature hierarchies for accurate object detection and semantic
  segmentation.
\newblock In \emph{CVPR}, pages 580--587, 2014.

\bibitem[Glorot et~al.(2011)Glorot, Bordes, and Bengio]{Glorot2011}
X.~Glorot, A.~Bordes, and Y.~Bengio.
\newblock Deep sparse rectifier neural networks.
\newblock In \emph{AISTATS}, pages 315--323, 2011.

\bibitem[Golowich et~al.(2018)Golowich, Rakhlin, and Shamir]{Golowich}
N.~Golowich, A.~Rakhlin, and O.~Shamir.
\newblock Size-independent sample complexity of neural networks.
\newblock In \emph{COLT}, volume~75, pages 297--299, 2018.

\bibitem[Graves et~al.(2013)Graves, Mohamed, and Hinton]{Graves2013}
A.~Graves, A.~Mohamed, and G.~Hinton.
\newblock Speech recognition with deep recurrent neural networks.
\newblock In \emph{ICASSP}, pages 6645--6649, 2013.

\bibitem[{Hebiri} and {Lederer}(2013)]{Hebiri2013Correlations}
M.~{Hebiri} and J.~{Lederer}.
\newblock How correlations influence lasso prediction.
\newblock \emph{IEEE Trans.\@ Inform.\@ Theory}, 59\penalty0 (3):\penalty0
  1846--1854, 2013.

\bibitem[Hebiri and Lederer(2020)]{Hebiri20}
M.~Hebiri and J.~Lederer.
\newblock Layer sparsity in neural networks.
\newblock \emph{arXiv:2006.15604}, 2020.

\bibitem[Hinton et~al.(2012)Hinton, Deng, Yu, Dahl, Mohamed, Jaitly, Senior,
  Vanhoucke, Nguyen, Sainath, and Kingsbury]{Hinton2012}
G.~Hinton, L.~Deng, D.~Yu, G.~Dahl, A.~Mohamed, N.~Jaitly, A.~Senior,
  V.~Vanhoucke, P.~Nguyen, T.~Sainath, and B.~Kingsbury.
\newblock Deep neural networks for acoustic modeling in speech recognition: The
  shared views of four research groups.
\newblock \emph{IEEE Signal Process.\@ Mag.\@}, 29\penalty0 (6):\penalty0
  82--97, 2012.

\bibitem[Jozefowicz et~al.(2016)Jozefowicz, Vinyals, Schuster, Shazeer, and
  Wu]{Jozefowicz2016}
R.~Jozefowicz, O.~Vinyals, M.~Schuster, N.~Shazeer, and Y.~Wu.
\newblock Exploring the limits of language modeling.
\newblock \emph{arXiv:1602.02410}, 2016.

\bibitem[Lederer(2010)]{lederer2010bounds}
J.~Lederer.
\newblock Bounds for {R}ademacher processes via chaining.
\newblock \emph{arXiv:1010.5626}, 2010.

\bibitem[Lederer(2020)]{Lederer20}
J.~Lederer.
\newblock No spurious local minima: on the optimization landscapes of wide and
  deep neural networks.
\newblock \emph{arXiv:2010.00885}, 2020.

\bibitem[Lederer and van~de Geer(2014)]{Lederer2014}
J.~Lederer and S.~van~de Geer.
\newblock New concentration inequalities for suprema of empirical processes.
\newblock \emph{Bernoulli}, 20\penalty0 (4):\penalty0 2020--2038, 2014.

\bibitem[Lederer and Vogt(2020)]{Lederer2020Estimating}
J.~Lederer and M.~Vogt.
\newblock Estimating the lasso's effective noise.
\newblock \emph{arXiv:2004.11554}, 2020.

\bibitem[Lederer et~al.(2019)Lederer, Yu, and Gaynanova]{Lederer2019}
J.~Lederer, L.~Yu, and I.~Gaynanova.
\newblock Oracle inequalities for high-dimensional prediction.
\newblock \emph{Bernoulli}, 25\penalty0 (2):\penalty0 1225--1255, 2019.

\bibitem[Ledoux and Talagrand(1991)]{Ledoux1991}
M.~Ledoux and M.~Talagrand.
\newblock \emph{Probability in {Banach} spaces: Isoperimetry and processes}.
\newblock Springer-Verlag Berlin, 1991.

\bibitem[Liu and Ye(2019)]{Liu2019}
H.~Liu and Y.~Ye.
\newblock High-dimensional learning under approximate sparsity: A unifying
  framework for nonsmooth learning and regularized neural networks.
\newblock \emph{arXiv:1903.00616}, 2019.

\bibitem[Long et~al.(2015)Long, Shelhamer, and Darrell]{Long2015}
J.~Long, E.~Shelhamer, and T.~Darrell.
\newblock Fully convolutional networks for semantic segmentation.
\newblock In \emph{CVPR}, pages 3431--3440, 2015.

\bibitem[Nair and Hinton(2010)]{Nair2010Rectified}
V.~Nair and G.~Hinton.
\newblock Rectified linear units improve restricted {Boltzmann} machines.
\newblock In \emph{ICLM}, pages 807--814, 2010.

\bibitem[Neyshabur et~al.(2014)Neyshabur, Tomioka, and Srebro]{Neyshabur2014}
B.~Neyshabur, R.~Tomioka, and N.~Srebro.
\newblock In search of the real inductive bias: on the role of implicit
  regularization in deep learning.
\newblock \emph{arXiv:1412.6614}, 2014.

\bibitem[Neyshabur et~al.(2015)Neyshabur, Tomioka, and Srebro]{Neyshabur2015}
B.~Neyshabur, R.~Tomioka, and N.~Srebro.
\newblock Norm-based capacity control in neural networks.
\newblock In \emph{COLT}, pages 1376--1401, 2015.

\bibitem[Ramachandran et~al.(2017)Ramachandran, Zoph, and
  Le]{Ramachandran2017Swish}
P.~Ramachandran, B.~Zoph, and Q.~Le.
\newblock Swish: A self-gated activation function.
\newblock \emph{arXiv:1710.05941}, 2017.

\bibitem[Scardapane et~al.(2017)Scardapane, Comminiello, Hussain, and
  Uncini]{Scardapane2017Group}
S.~Scardapane, D.~Comminiello, A.~Hussain, and A.~Uncini.
\newblock Group sparse regularization for deep neural networks.
\newblock \emph{Neurocomputing}, 241:\penalty0 81--89, 2017.

\bibitem[Schmidt-Hieber(2017)]{Hieber2017}
J.~Schmidt-Hieber.
\newblock Nonparametric regression using deep neural networks with {ReLU}
  activation function.
\newblock \emph{arXiv:1708.06633}, 2017.

\bibitem[Szegedy et~al.(2015)Szegedy, Liu, Jia, Sermanet, Reed, Anguelov,
  Erhan, Vanhoucke, and Rabinovich]{Szegedy2015}
C.~Szegedy, W.~Liu, Y.~Jia, P.~Sermanet, S.~Reed, D.~Anguelov, D.~Erhan,
  V.~Vanhoucke, and A.~Rabinovich.
\newblock Going deeper with convolutions.
\newblock In \emph{CVPR}, pages 1--9, 2015.

\bibitem[Tibshirani(1996)]{lasso}
R.~Tibshirani.
\newblock Regression shrinkage and selection via the lasso.
\newblock \emph{J.\@ R.\@ Stat.\@ Soc.\@ Ser.\@ B.\@ Stat.\@ Methodol.\@},
  58\penalty0 (1):\penalty0 267--288, 1996.

\bibitem[van~de Geer(2000)]{Sara2000}
S.~van~de Geer.
\newblock \emph{Empirical processes in M-estimation}.
\newblock Cambridge Univ.\@ Press, 2000.

\bibitem[van~de Geer(2016)]{van2016estimation}
S.~van~de Geer.
\newblock \emph{Estimation and testing under sparsity}.
\newblock Springer, 2016.

\bibitem[van~de Geer and Lederer(2013)]{Sara2013Orliz}
S.~van~de Geer and J.~Lederer.
\newblock The {Bernstein–Orlicz} norm and deviation inequalities.
\newblock \emph{Probab.\@ Theory Related Fields}, 157:\penalty0 225--250.,
  2013.

\bibitem[van~der Vaart and Wellner(1996)]{Vaart1996}
A.~van~der Vaart and J.~Wellner.
\newblock \emph{Weak convergence and empirical processes}.
\newblock Springer, 1996.

\bibitem[Vershynin(2018)]{Vershynin2018}
R.~Vershynin.
\newblock \emph{High-dimensional probability: An introduction with applications
  in data science}.
\newblock Cambridge Univ.\@ Press, 2018.

\bibitem[Yarotsky(2017)]{Yarotsky2017}
D.~Yarotsky.
\newblock Error bounds for approximations with deep {ReLU} networks.
\newblock \emph{Neural Networks}, 94:\penalty0 103--114, 2017.

\bibitem[Zhang et~al.(2016)Zhang, Lee, and Jordan]{Zhang2016L1}
Y.~Zhang, J.~Lee, and M.~Jordan.
\newblock $\ell_1$-regularized neural networks are improperly learnable in
  polynomial time.
\newblock In \emph{ICML}, pages 993--1001, 2016.

\end{thebibliography}
\end{document}